\documentclass[preprint,12pt,authoryear]{elsarticle}

\usepackage{amsmath,amsfonts}
\usepackage{array}
\usepackage[caption=false,font=normalsize,labelfont=sf,textfont=sf]{subfig}
\usepackage{textcomp}
\usepackage{stfloats}
\usepackage{url}
\usepackage{verbatim}
\usepackage{graphicx}
\usepackage{cite}
\usepackage{dsfont}
\usepackage[linesnumbered,ruled,vlined]{algorithm2e}
\usepackage{soul}
\usepackage{xcolor}
\usepackage[a4paper, total={6in, 8in}]{geometry}

\usepackage{pdflscape}
\newcommand{\bb}[1]{\mathbf{#1}}
\newcommand{\bbb}[1]{\boldsymbol{#1}}

\usepackage[hidelinks]{hyperref}
\usepackage{mathtools,leftidx}
\usepackage{amssymb}
\begin{document}
	\begin{frontmatter}
	
	\title{An Improved Multi-State Constraint Kalman Filter for Visual-Inertial Odometry}
	\author[one]{M.R. Abdollahi$^{\dagger, }$}
	\author[one]{Seid H. Pourtakdousti}
	\author[two]{M.H. Yoosefian Nooshabadi$^{\dagger, }$}
	\author[two]{H.N. Pishkenari}
	
%	\author{M.R. Abdollahi,
%		\autho Seid H. Pourtakdoust, M.H. Yoosefian Nooshabadi and H.N. Pishkenari}
	%%% use optional labels to link authors explicitly to addresses:
	 \affiliation[one]{organization={Department of Aerospace Engineering},
		             addressline={Sharif University of Technology},
		             city={Tehran},
		             country={Iran}}
	   
	    \affiliation[two]{organization={Department of Mechanical Engineering},
	   	addressline={Sharif University of Technology},
	   	city={Tehran},
	   	country={Iran}}
   	
%   	 \affiliation[three]{organization={sharif},
%   		addressline={},
%   		city={},
%   		postcode={},
%   		state={},
%   		country={}}
%   	
%   	 \affiliation[four]{organization={sharif},
%   		addressline={},
%   		city={},
%   		postcode={},
%   		state={},
%   		country={}}

	%%%
	%%% \affiliation[label2]{organization={},
		%%%             addressline={},
		%%%             city={},
		%%%             postcode={},
		%%%             state={},
		%%%             country={}}

	%\IEEEpubid{0000--0000/00\$00.00~\copyright~2023 IEEE}
	
%	\maketitle
	
	\begin{abstract}
		Fast pose estimation (PE) plays a vital role for agile autonomous robots in successfully carrying out their tasks. While Global Navigation Satellite Systems (GNSS) such as Global Positinoing System (GPS) have been traditionally used along with Inertial Navigation Systems (INS) for PE, their viability is compromised in indoor and urban environments due to their low update rates and inadequate signal coverage. Visual-Inertial Odometry (VIO) is gaining popularity as a practical alternative to GNSS/INS systems in GNSS-denied environments. Among various VIO-based methods, the Multi-State Constraint Kalman Filter (MSCKF) has garnered significant attention due to its robustness, speed and accuracy. Nevertheless, high computational cost of image processing is still challenging for real-time implementation on resource-constrained vehicles. In this paper, an enhanced version of the MSCKF is proposed. The proposed approach differs from the original MSCKF in the feature marginalization and state pruning steps of the algorithm. This new design results in a faster algorithm with comparable accuracy. We validate the proposed algorithm using both an open-source dataset and real-world experiments. It is demonstrated that the proposed Fast-MSCKF (referred to as FMSCKF) is approximately six times faster and at least 20\% more accurate in final position estimation compared to the standard MSCKF. 
	\end{abstract}

\end{frontmatter}

		\def\thefootnote{$\dagger$}\footnotetext{These authors contributed equally to this work.}\def\thefootnote{\arabic{footnote}}
		
	\begin{frame}
		video link: \href{https://www.youtube.com/watch?v=4RaHqA068zQ}{experimental test}
	\end{frame}
	
\section{Introduction}\label{section1}
In recent years, the utilization of quadrotors has witnessed a significant upsurge across various applications \citep{ref1,ref2,ref3}. Accurate and reliable pose estimation serves as a crucial step in designing guidance and control systems for quadrotors. GPS and IMU data fusion has been widely used for pose estimation in autonomous vehicles \citep{ref4,ref34,ref35,ref36}. However, this method has some drawbacks in certain scenarios. For instance, satellite signals may not be readily available in various settings, such as forests, indoor areas, and urban environments with tall buildings. Moreover, the update rate of many satellite receivers is notably low and with a slight delay. As a result, alternative approaches have been explored to replace GNSS/INS systems and address their limitations. Although it does not provide absolute pose (i.e., the pose of the system with respect to an Earth-fixed frame), camera-IMU fusion is one of the emerging methods for relative PE, due to the low cost and high quality of the sensors. This approach is commonly referred to as Visual Inertial Odometry (VIO) in the relevant literature. VIO approaches are considered safe, cost-effective and robust replacements for GNSS/INS systems in applications where it suffices to have relative PE.

There are various methods for VIO in the literature, including ORB-SLAM \citep{ref5}, SVO+GTSAM \citep{ref6}, CNN-SVO \citep{ref7}, DSO \citep{ref8}, VINS-Mono \citep{ref9}, and Kimera \citep{ref10}. VIO methods can be categorized into two major groups, namely, \textit{filter-based} and \textit{optimization-based} approaches. Optimization-based methods tend to be more computationally expensive, which makes them suitable for offline and post-processing applications. On the other hand, the low computational load and relatively high accuracy of filter-based methods renders them well-suited for real-time implementations \citep{ref11}. Given that this paper's focus is on fast PE, we opt for the latter approach.
In certain filter-based methods, landmark positions are stored in the state vector (see ROVIO \citep{ref12}, EKF- and UKF- SLAM \citep{ref13}). However, as the number of landmark positions increases, the size of the state vector grows proportionally, resulting in a considerable rise in computational demands \citep{ref14}. Therefore, for real-time applications, only a limited number of landmark positions can be used. On the other hand, some other filter-based approaches, such as the MSCKF, store a number of camera poses in the state vector.

The Multi-State Constraint Kalman Filter (MSCKF) was originally developed by Mourikis and Roumeliotis in their seminal paper \citep{ref15}. The MSCKF uses an Error-State Extended Kalman filter to fuse IMU and camera data and, unlike other KF-based methods, does not store the positions of landmarks in the state vector. Instead, a series of recent positions and orientations (i.e., poses) of the camera are kept in the state vector. This alternative strategy significantly reduces the computational cost while enhancing estimation accuracy and robustness.

In general, to improve the performance of a VIO algorithm, three main characteristics need to be considered, namely, \textit{accuracy}, \textit{robustness}, and \textit{output rate}. There are many studies in the literature that attempt to improve these three features. To improve accuracy, a Patch-based MSCKF \citep{ref16}, has been developed. Unlike the original MSCKF algorithm, in which image features are utilized, in the patch-based MSCKF, a direct method is employed. In this method, the light intensities of pixels in distinct patches are used for estimation. This approach achieves, on average, 23\% enhancement in accuracy compared to the original MSCKF \citep{ref16}.

The Local-Optimal MSCKF (LOMSCKF) has been developed in an attempt to improve estimation accuracy using nonlinear optimization \citep{ref6,ref17}. The LOMSCKF employs pre-integrated IMU and camera measurements. Moreover, Stereo MSCKF (S-MSCKF), which uses a stereo camera instead of a single camera, has been proposed to enhance the performance of the original MSCKF. The S-MSCKF has shown more robustness with a modest increase in computational cost \citep{ref18}.

One limitation of EKF-based methods lies in the inherent unobservability of the yaw angle, rendering them intrinsically inconsistent. Huang has considered this problem and proposed potential solutions, including Observability-Constrained EKF (OC-EKF) \citep{ref19} and First-Estimates Jacobian (FEJ-EKF) \citep{ref20}. Furthermore, MA has introduced Ackermann MSCKF (ACK-MSCKF) algorithm \citep{ref21}, in which the MSCKF has been modified to be used for ground vehicles. In this work, the unobservable states of ground robots are investigated, and an algorithm is presented to resolve the unobservability problem.

There are also numerous recent works focusing on the integration of machine learning techniques into visual-inertial navigation to enhance robustness and accuracy \citep{ref22,ref23}. For instance, a convolutional neural network has been implemented in the update step of the MSCKF algorithm, which has resulted in higher robustness compared to the original MSCKF \citep{ref24}. However, learning-based approaches often lack formal convergence guarantees, which makes them unreliable for safety-critical application.

As discussed above, numerous studies have been conducted with the goal of enhancing the original MSCKF algorithm. The majority of these endeavors have aimed to improve the algorithm's accuracy or address the observability issue associated with the yaw angle. However, relatively few studies have tried to decrease the computational cost of the algorithm. Yet, the high computational cost remains the primary challenge for resource-constrained robots. Therefore, reducing this computational cost is of great importance. To this end, this paper attempts to reduce the computational cost of the MSCKF algorithm using a new feature management method.

The rest of the paper is organized as follows. The next section covers the formulation of the MSCKF. Section \ref{section3} constitutes the core contribution of this work, elucidating the feature management methodology of both the original MSCKF and the proposed FMSCKF. The implementation results obtained from both the MSCKF and FMSCKF on an open-source dataset and in real-world experiments are presented in section \ref{section4}. In section \ref{section5}, a detailed discussion of the results is provided. Finally, conclusions are provided in section \ref{section6}.

\section{The Multi State Constraint Kalman Filter (MSCKF)}\label{section2}
The MSCKF is an error-state extended Kalman filter that can be used to estimate the position and orientation of a camera-IMU system. More precisely, the algorithm estimates the pose of the frame attached to the IMU, $\left\{I\right\}$, with respect to a global (reference) frame, $\left\{G\right\}$. The equations in this section are based on \citep{ref15}. The state vector of the filter consists of two parts. The first part, which evolves during the propagation step, is the IMU state
\begin{equation}\label{eq: state IMU}
	\mathbf{X}_I = 
	\begin{bmatrix}
		\prescript{I}{}{\bb{q}}_G^\top & \mathbf{b}_g^\top & \prescript{G}{}{\mathbf{v}}_I^\top & \mathbf{b}_a^\top & \prescript{G}{}{\mathbf{p}}_I^\top
	\end{bmatrix}^\top,
\end{equation}
in which $\prescript{I}{}{\bb{q}}_G \in \mathbb{R}^4$ is the unit quaternion indicating the rotation from the frame $\left\lbrace G \right\rbrace$ to the frame $\left\lbrace I \right\rbrace$, $\prescript{G}{}{\bb{v}}_I \in \mathbb{R}^3$ and $\prescript{G}{}{\mathbf{p}}_I\in \mathbb{R}^3$ 
are the velocity and position vectors of the frame $\left\lbrace I \right\rbrace$ with respect to and expressed in the frame $\left\lbrace G \right\rbrace$, and $\mathbf{b}_g\in \mathbb{R}^3$ and $\mathbf{b}_a \in \mathbb{R}^3$ are gyroscope and accelerometer biases, respectively. The second part of the state vector includes a number of poses of the camera frames $\left\lbrace C \right\rbrace$  with respect to the global frame $\left\lbrace G \right\rbrace$. This part is added to the state vector during the augmentation step (section \ref{section2b}). The algorithm consists of three steps, which are explained in the following sections.

\subsection{The Propagation Step}
Every time a new IMU measurement is received, the first part of the state vector and its corresponding covariance matrix are propagated. Considering a calibrated IMU (i.e., misalignment, temperature effects, scale-factor, etc. are accounted for), the measurement models for the gyroscope and the accelerometer are
\begin{equation}\label{eq: gyro measurement}
	\bbb{\omega}_m=\bbb{\omega}+\bb{b}_g+\bb{n}_g,
\end{equation}
\begin{equation}\label{eq: accel measurement}
	\bb{a}_m = \prescript{I}{}{\bb{R}}_G \left( \prescript{G}{}{\bb{a}} - \prescript{G}{}{\bb{g}} \right) + \bb{b}_a + \bb{n}_a,
\end{equation}
in which $\bbb{\omega}_m \in \mathbb{R}^3$ and $\bbb{\omega} \in \mathbb{R}^3$ are the measured and true angular velocities, $\bb{a}_m \in \mathbb{R}^3$ is the measured body acceleration, $\prescript{I}{}{\bb{R}}_G \in \mathbb{R}^{3\times3}$ is the rotation matrix corresponding to the quaternion $\prescript{I}{}{\bb{q}}_G$, $\bb{n}_g \in \mathbb{R}^3$ and $\bb{n}_a \in \mathbb{R}^3$ are zero mean	 white process noise vectors, $\prescript{G}{}{\bb{a}} \in \mathbb{R}^3$ is the acceleration of the frame $\left\lbrace I \right\rbrace$ expressed in $\left\lbrace G \right\rbrace$, and finally, $\prescript{G}{}{\bb{g}} \in \mathbb{R}^3$ is the local gravity vector expressed in $\left\lbrace G \right\rbrace$. Bias vectors are modeled as random walk processes, hence, the time derivatives of entries of the IMU state vector is
\begin{equation}\label{eq: quaternion ODE}
	%	\cite{msckf2}
	\begin{split}
		\prescript{I}{}{\dot{\bb{q}}}_{G}({t}) &=\frac{1}{2} \bbb{\Omega}\left(\prescript{I}{}{\bbb{\omega}}(t)\right)\prescript{I}{}{\bb{q}}_{G}({t}),
	\end{split}
\end{equation}
\begin{equation}\label{eq: position ODE}
	\begin{split}
		\prescript{G}{}{\dot{\bb{p}}}_I({t}) &= \prescript{G}{}{\bb{v}}_I({t}),
	\end{split}
\end{equation}
\begin{equation}\label{eq: velocity ODE}
	\begin{split}
		\prescript{G}{}{\dot{\bb{v}}}_I({t}) &= \prescript{G}{}{\bb{a}}_I({t}),
	\end{split}
\end{equation}
\begin{equation}\label{eq: gyro bias ODE}
	\begin{split}
		{\dot{\bb{b}}}_g({t}) &= \bb{n}_{b_g}({t}),
	\end{split}
\end{equation}
\begin{equation}\label{eq: accel bias ODE}
	\begin{split}
		{\dot{\bb{b}}}_a(t) &= \bb{n}_{b_a}(t).
	\end{split}
\end{equation}
In \eqref{eq: gyro bias ODE} and \eqref{eq: accel bias ODE}, $\bb{n}_{b_{g}}\in \mathbb{R}^3$ and $\bb{n}_{b_{a}}\in \mathbb{R}^3$ are random walk noise vectors and
\begin{equation}\label{eq: Omega def}
	%	\cite{msckf2}
	\bbb{\Omega}(\bbb{\omega})=
	\begin{bmatrix}
		-[\bbb{\omega} \times] & \bbb{\omega} \\
		-\bbb{\omega}^\top & \bb{0}
	\end{bmatrix}, \quad
	[\bbb{\omega} \times] = \begin{bmatrix}
		0 & -\omega_z & \omega_y \\
		\omega_z & 0 & -\omega_x \\
		\omega_y & \omega_x & 0
	\end{bmatrix}.
\end{equation}
Applying the expectation operator to equations \eqref{eq: quaternion ODE} to \eqref{eq: accel bias ODE} yields
\begin{equation}\label{eq: quaternion ODE expectation}
	%	\cite{msckf2}
	\begin{split}
		\prescript{I}{}{\dot{\hat{\bb{q}}}}_G &=\frac{1}{2} \bbb{\Omega}\left(\bbb{\hat{\omega}}\right)\prescript{I}{}{\hat{\bb{q}}}_G,
	\end{split}
\end{equation}
\begin{equation}\label{eq: position ODE expectation}
	\begin{split}
		{\prescript{G}{}{\dot{\hat{\bb{p}}}}}_I &= \prescript{G}{}{\hat{\bb{v}}}_I,
	\end{split}
\end{equation}
\begin{equation}\label{eq: velocity ODE expectation}
	\begin{split}
		{\prescript{G}{}{\dot{\bb{\hat{\bb{v}}}}}}_I &= \prescript{I}{}{\bb{R}}^\top_{G}  \bb{\hat{a}} + \prescript{G}{}{\bb{g}},
	\end{split}
\end{equation}
\begin{equation}\label{eq: gyro bias ODE expectation}
	\begin{split}
		{\dot{\bb{\hat{\bb{b}}}}}_g &= \bb{0},
	\end{split}
\end{equation}
\begin{equation}\label{eq: accel bias ODE expectation}
	\begin{split}
		{\dot{\hat{\bb{b}}}}_a &= \bb{0},
	\end{split}
\end{equation}
where $\bbb{\hat{\omega}} = \bbb{\omega}_m - \bb{\hat{b}}_g$ and $\bb{\hat{a}} = \bb{a}_m - \bb{\hat{b}}_a$. These equations can be solved either by continuous-time integration (using differential equations) or discrete-time integration (using methods such as fourth order Runge-Kutta) to obtain the IMU state propagation equations. The covariance matrix is also propagated. This matrix, at time step $k+1$, can be partitioned as
\begin{equation}\label{eq: cov matrix partition}
	%	\cite{msckf2}
	\bb{P}=\begin{bmatrix}
		\bb{P}_{{II}_{{k+1|k}}} & \bb{P}_{IC_{{k+1|k}}} \\
		\bb{P}_{{CI}_{{k+1|k}}} & \bb{P}_{{CC}_{{k|k}}}
	\end{bmatrix},
\end{equation}
in which the $\bb{P}_{{II}_{{k+1|k}}}$ is the covariance matrix related to the IMU state (the correlation between IMU state variables) and can be calculated through numerical integration of the following Lyapunov equation
\begin{equation}\label{eq: Lyapunov equation for PII}
	\begin{split}
		\dot{\bb{P}}_{II} & = \bb{F}\bb{P}_{{II}} + \bb{P}_{II}\bb{F}^\top + \bb{G}\bb{Q}_I\bb{G}^\top,
	\end{split}
\end{equation}
in which
\begin{equation}\label{eq: F matrix}
	%	\cite{msckf2}
	\begin{split}
		\bb{F} & = \begin{bmatrix}
			-\left[\prescript{{I}}{}{\hat{\bbb{\omega}}}\times\right] & -\bb{I}_3 & \bb{0}_3 & \bb{0}_3 & \bb{0}_3 \\
			\bb{0}_3 & \bb{0}_3 & \bb{0}_3 & \bb{0}_3 & \bb{0}_3 \\
			-\prescript{G}{}{\hat{\bb{R}}}_I\left[\bb{\hat{a}}\times\right] & \bb{0}_3 & \bb{0}_3 & -\prescript{G}{}{\hat{\bb{R}}}_I & \bb{0}_3 \\
			\bb{0}_3 & \bb{0}_3 & \bb{0}_3 & \bb{0}_3 & \bb{0}_3 \\
			\bb{0}_3 & \bb{0}_3 & \bb{I}_3 & \bb{0}_3 & \bb{0}_3 \\
		\end{bmatrix} ,
	\end{split}
\end{equation}
\begin{equation}\label{eq: G matrix}
	%	\cite{msckf2}
	\begin{split}
		\bb{G} & = \begin{bmatrix}
			-\bb{I}_3 & \bb{0}_3 & \bb{0}_3 & \bb{0}_3 \\
			\bb{0}_3 & -\bb{I}_3 & \bb{0}_3 & \bb{0}_3 \\
			\bb{0}_3 & \bb{0}_3 & -\prescript{G}{}{\hat{\bb{R}}}_I & \bb{0}_3 \\
			\bb{0}_3 & \bb{0}_3 & \bb{0}_3 & -\bb{I}_3 \\
			\bb{0}_3 & \bb{0}_3 & \bb{0}_3 & \bb{0}_3
		\end{bmatrix},
	\end{split}
\end{equation}
where $\prescript{G}{}{\hat{\bb{R}}}_I$ and $\left[\bb{\hat{a}}\times\right]$ are clear from the context, and $\bb{Q}_I$ is the covariance matrix corresponding to the noise vector
\begin{equation}\label{eq: noise vector}
	\bb{n}_I =\begin{bmatrix}
		\bb{n}^\top_{{g}} & \bb{n}^\top_{b_g}  & \bb{n}^\top_{{a}}  &  \bb{n}^\top_{b_a} 
	\end{bmatrix}^\top.
\end{equation}

The next term in the covariance matrix in \eqref{eq: cov matrix partition} is $\bb{P}_{{IC}_{{k+1|k}}}$, which represents the correlation between the IMU and the camera states and is propagated using
\begin{equation}\label{eq: equation for PIC}
	\bb{P}_{{IC}_{{k+1|k}}} = \bbb{\Phi}\left(t_k + T, t_k\right) \bb{P}_{{IC}_{{k|k}}},
\end{equation}
where $T$ is the sampling time of the IMU, and $\bbb{\Phi}$ is the state transition matrix that evolves over time according to the following equation
\begin{equation}\label{eq: state transition ODE}
	\dot{\bbb{\Phi}}\left(t_k + \tau, t_k\right) = \bb{F}\bbb{\Phi}\left(t_k + \tau, t_k\right),
\end{equation}
with $\bb{F}$ given in \eqref{eq: F matrix}.

\subsection{The Augmentation Step}\label{section2b}
Upon recording an image, the camera pose is calculated based on the most recent estimate of the IMU pose. Assume that the filter is currently at the $\left(k+1\right)$-th time step, and the state vector already includes $l$ camera poses. When the $\left(l+1\right)$-th image is received, its pose is calculated using
\begin{equation}\label{eq: quaternion augmentation}
	%	\cite{msckf2}
	\prescript{C_{l+1}}{}{\hat{\bb{q}}}_G= \prescript{C}{}{\bb{q}}_I \bbb{\otimes} \prescript{I_{k+1}}{}{\hat{\bb{q}}}_G,
\end{equation}
\begin{equation}\label{eq: position augmentation}
	%	\cite{msckf2}
	\prescript{G}{}{\hat{\bb{p}}}_{C_{l+1}} = \prescript{G}{}{\hat{\bb{p}}}_{I_{k+1}} + \prescript{G}{}{\hat{\bb{R}}}_{I_{k+1}} \prescript{I}{}{\bb{p}}_C,
\end{equation}
in which $\prescript{I}{}{\bb{p}_C}$ and $\prescript{C}{}{\bb{q}_I}$ are the translation vector and the quaternion between the IMU frame and the camera frame, respectively. Since the two sensors are rigidly attached to each other, $\prescript{I}{}{\bb{p}_C}$ and $\prescript{C}{}{\bb{q}_I}$ are two constants, and can be computed offline. In the next step, the state vector is augmented according to
\begin{equation}
	%	\cite{msckf2}
	\bb{\widehat{X}}_{aug} = \begin{bmatrix}
		\bb{\hat{X}}^\top & \prescript{{C}_{{l+1}}}{}{\hat{\bb{q}}}_G^\top & \prescript{{G}}{}{\hat{\bb{p}}}_{{C_{l+1}}}^\top
	\end{bmatrix}^\top,
\end{equation}
in which $\bb{\widehat{X}}$ is the state vector prior to augmentation. In addition to the state vector, the covariance matrix should be augmented using
\begin{equation}\label{eq: augmented cov matrix}
	%	\cite{msckf2}
	\bb{P} = \begin{bmatrix}
		\bb{I}_{6l+15} \\
		\bb{J}
	\end{bmatrix}
	\bb{P}
	\begin{bmatrix}
		\bb{I}_{6l+15} \\
		\bb{J}
	\end{bmatrix}^\top,
\end{equation}
in which $\bb{J}$ is the Jacobian matrix, that is calculated using
\begin{equation}\label{eq: Jacobian matrix}
	%	\cite{msckf2}
	\bb{J}=\begin{bmatrix}
		\prescript{C}{}{\bb{R}}_I & \bb{0}_{3\times 9} & \bb{0}_3 & \bb{0}_{3\times 6l}\\
		\left[\left(\prescript{G}{}{\hat{\bb{R}}}_{I_{k+1}} \prescript{I}{}{\bb{p}}_C\right) \times\right] & \bb{0}_{3\times 9} & \bb{I}_3 & \bb{0}_{3\times 6l}
	\end{bmatrix}.
\end{equation}

\subsection{The Update Step}
The update step is carried out using features extracted from images, the poses of which are available in the state vector. The policy based on which these features are chosen is the subject of section \ref{section3}, and is the key distinction between the proposed method and the original MSCKF. Assume $m$ features are observed in $n_j$ images $\left(l-n_j, \cdots, l-1\right)$ with $l$ being the index of the most recent recorded image. Each image contains pixel coordinates corresponding to different features. Therefore, the camera measurement model for the $j$-th $\left(j = 1, \cdots , m\right)$ feature observed in the $i$-th $\left(i = 1, \cdots , n_j\right)$ image is
\begin{equation}\label{eq: measurement model}
	\prescript{{C}_{i}}{}{\bb{z}}_{{f}_{j}} = \frac{1}{\prescript{C_i}{}{{Z}}_{f_j}}
	\begin{bmatrix}
		\prescript{C_i}{}{{X}}_{f_j}\\
		\prescript{C_i}{}{{Y}}_{f_j}
	\end{bmatrix} + \prescript{C_i}{}{\bb{n}}_{f_j},
\end{equation}
where $\prescript{{C}_i}{}{\bb{n}}_{f_j} \in \mathbb{R}^2$ is the measurement noise with the covariance $\prescript{C_i}{}{\bb{R}}_{f_j} = \sigma^{2}_{im}\bb{I}_2$, and
$
\prescript{C_i}{}{\bb{p}}_{f_j} = 
\begin{bmatrix}
	\prescript{C_i}{}{X}_{f_j} & \prescript{C_i}{}{Y}_{f_j} & \prescript{C_i}{}{Z}_{f_j}
\end{bmatrix}^\top
$ is the position of the $j$-th feature in the $i$-th camera frame. Utilizing all $n_j$ measurements of the feature, it is possible to estimate the position of the feature in the $\left\{G\right\}$ frame via triangulation (see \citep{ref25,ref26}). Assume that the estimated position of the $j$-th feature in the global frame is denoted as $ \prescript{G}{}{\bb{\hat{p}}}_{f_j} $. The estimate of the position of the $j$-th feature in the $i$-th image frame is calculated via

\begin{equation}\label{eq: estimate feature position}
	\prescript{C_i}{}{\hat{\bb{p}}}_{f_j} = \prescript{C_i}{}{\hat{\bb{R}}}_G\left(\prescript{G}{}{\bb{\hat{p}}}_{f_j} - \prescript{G}{}{\bb{\hat{p}}}_{C_i}\right) = 
	\begin{bmatrix}
		\prescript{C_i}{}{{\hat{X}}}_{f_j}\\
		\prescript{C_i}{}{{\hat{Y}}}_{f_j}\\
		\prescript{C_i}{}{{\hat{Z}}}_{f_j}
	\end{bmatrix}.
\end{equation}
The residual is calculated using the measurement of the feature
\begin{equation}\label{eq: residual}
	\prescript{C_i}{}{\bb{r}}_{f_j} = \prescript{C_i}{}{\bb{z}}_{f_j} - \prescript{C_i}{}{\bb{\hat{z}}}_{f_j},  
\end{equation}
in which
\begin{equation}\label{eq: estimate of z}
	\prescript{C_i}{}{\bb{\hat{z}}}_{f_j} = \frac{1}{\prescript{C_i}{}{{\hat{Z}}}_{f_j}}
	\begin{bmatrix}
		\prescript{C_i}{}{{\hat{X}}}_{f_j} & \prescript{C_i}{}{{\hat{Y}}}_{f_j}
	\end{bmatrix}^\top.
\end{equation}

Subsequently, the Jacobians of the measurement model with respect to the position of the feature, $\prescript{C_i}{}{\bb{H}}_{f_j}$, and with respect to the state vector, $\prescript{C_i}{}{\bb{H}}_{X_j}$, are calculated using
\begin{equation}\label{eq: Jacobian wrt feature}
	\prescript{{C}_i}{}{\bb{H}}_{{f}_j} = \prescript{{C}_i}{}{\bb{J}}_{{f}_j} \prescript{{C}_i}{}{\hat{\bb{R}}}_{{G}},
\end{equation}
\begin{equation}\label{eq: Jacobian wrt state}
	\prescript{{C}_i}{}{\bb{H}}_{{X}_j} = 
	\begin{bmatrix}
		\bb{0}_{{2\times15}} & \cdots & \prescript{{C}_i}{}{\bb{H}}_{f_j}
		\begin{bmatrix}
			\left(\prescript{G}{}{\hat{\bb{p}}}_{{f}_j} - \prescript{G}{}{\hat{\bb{p}}}_{{C}_i}\right)\times
		\end{bmatrix} & -\prescript{{C}_i}{}{\bb{H}}_{{f}_j}
	\end{bmatrix} ,
\end{equation}
where
\begin{equation}\label{eq: Jacobian for Hf}
	\prescript{C_i}{}{\bb{J}}_{f_j} = \left(\frac{1}{\prescript{C_i}{}{{\hat{Z}}}_{f_j}}\right)^{2}
	\begin{bmatrix}
		\prescript{C_i}{}{{\hat{Z}}}_{f_j} & 0 & -\prescript{C_i}{}{{\hat{X}}}_{f_j}\\
		0 & \prescript{C_i}{}{{\hat{Z}}}_{f_j} & -\prescript{C_i}{}{{\hat{Y}}}_{f_j}
	\end{bmatrix}.
\end{equation}
These calculations are performed for all $n_j$ images in which the $j$-th feature has been observed. Concatenating all calculations for the $j$-th feature yields
\begin{equation}\label{eq: complete residual}
	\bb{r}_{{f}_j}= 
	\begin{bmatrix}
		\prescript{C_1}{}{\bb{r}}_{f_j}^\top & \cdots & \prescript{C_{n_j}}{}{\bb{r}}_{f_j}^\top
	\end{bmatrix}^\top,
\end{equation}
\begin{equation}\label{eq: complete Hf}
	\bb{H}_{{f}_j}= 
	\begin{bmatrix}
		\prescript{C_1}{}{\bb{H}}_{f_j}^\top & \cdots & \prescript{C_{n_j}}{}{\bb{H}}_{f_j}^\top
	\end{bmatrix}^\top,
\end{equation}
\begin{equation}\label{eq: complete Hx}
	\bb{H}_{X_j}= 
	\begin{bmatrix}
		\prescript{C_1}{}{\bb{H}}_{X_j}^\top & \cdots & \prescript{C_{n_j}}{}{\bb{H}}_{X_j}^\top
	\end{bmatrix}^\top.
\end{equation}

Let $\bb{A}_{f_j}$ be the left null-space of $\bb{H}_{f_j}$. Define
\begin{equation}\label{eq: equation for Hoj}
	\bb{H}_{o_j}= \bb{A}_{f_j}^\top\bb{H}_{X_j},
\end{equation}
\begin{equation}\label{eq: equation for Roj}
	\bb{R}_{o_j} = \sigma_{im}^2\bb{A}_{f_j}^\top\bb{A}_{f_j},
\end{equation}
\begin{equation}\label{eq: equation for roj}
	\bb{r}_{o_j} = \bb{A}_{f_j}^\top \bb{r}_{f_j}.
\end{equation}
Matrices $\bb{H}_{o_j}$, $\bb{R}_{o_j} $, and $\bb{r}_{o_j}$ are calculated for all $m$ features, which are to be used to carry out the update step. Stacking the calculated matrices for all $m$ features yields
\begin{equation}\label{eq: equation for Ho}
	\bb{H}_o = 
	\begin{bmatrix}
		\bb{H}_{o_1}^\top & \cdots & \bb{H}_{o_m}^\top
	\end{bmatrix}^\top, 
\end{equation}
\begin{equation}\label{eq: equation for ro}
	\bb{r}_o = 
	\begin{bmatrix}
		\bb{r}_{o_1}^\top & \cdots & \bb{r}_{o_m}^\top
	\end{bmatrix}^\top, 
\end{equation}
\begin{equation}\label{eq: equation for Ro}
	\bb{R}_{o} = \text{diag}\left(\bb{R}_{o_1}, \dots, \bb{R}_{o_m}\right),
\end{equation}
where $\bb{r}_o \in \mathbb{R}^d$ is the residual computed for all ${m}$ features observed in ${n}_j$ images, and 
\begin{equation*}
	%	\label{deqn_ex1a}
	d = \sum_{j=1}^{m} \left(2n_j -3\right).
\end{equation*}

Since the number of features can be large, an additional step can be taken to reduce the computational cost. Specifically, reduced QR-decomposition can be utilized to decompose the $\bb{H}_o$ matrix as
\begin{equation}\label{eq: decomposition of Ho}
	\bb{H}_o = 
	\begin{bmatrix}
		\bb{Q}_1 & \bb{Q}_2
	\end{bmatrix}
	\begin{bmatrix}
		\bb{T}_H\\
		\bb{0}
	\end{bmatrix},
\end{equation}
where $\bb{Q}_1$ and $\bb{Q}_2$ are unitary matrices, and $\bb{T}_H$ is an upper triangular matrix. As a result, the matrices calculated using \eqref{eq: equation for Ho} to \eqref{eq: equation for Ro} can be reduced to
\begin{equation}\label{eq: equation for Hn}
	\bb{H}_n = \bb{T}_H,
\end{equation}
\begin{equation}\label{eq: equation for rn}
	\bb{r}_n = \bb{Q}_1^\top \bb{r}_o,
\end{equation}
\begin{equation}\label{eq: equation for Rn}
	\bb{R}_n = \bb{Q}_1^\top \bb{R}_o \bb{Q}_1.
\end{equation}

Next, the Kalman gain is computed via
\begin{equation}\label{eq: Kalman gain}
	\bb{K} = \bb{P}_{k+1|k} \bb{H}_{n}^\top \bb{S}^{-1},
\end{equation}
where 
\begin{equation}\label{eq: equation for S in kalman gain}
	\bb{S} = \bb{H}_n \bb{P}_{k+1|k} \bb{H}_{n}^\top.
\end{equation}
Finally, the Kalman gain in \eqref{eq: Kalman gain} is used to update the covariance matrix and the state vector according to
\begin{equation}\label{eq: cov matrix update}
	\bb{P}_{k+1|k+1} = \bb{P}_{k+1|k} - \bb{KSK}^\top,
\end{equation}
\begin{equation}\label{eq: error state update}
	\widetilde{\bb{X}} = \bb{Kr}_{n},
\end{equation}
\begin{equation}\label{eq: state update}
	\widehat{\bb{X}}_{k+1|k+1} = \widehat{\bb{X}}_{k+1|k} \bbb{\oplus} \widetilde{\bb{X}},
\end{equation}
where the $\oplus$ represents the general form of addition, which for the quaternion entry of the state vector is equivalent to the quaternion multiplication, and for other entries is the same as normal addition. The updated state vector and covariance matrix will be used in the next propagation step of the subsequent iteration of the filter. 

Note that since the MSCKF is an Error-State Extended Kalman Filter, the orientation error-state is minimal, as we define
\begin{equation*}
	\delta \bb{q} = \begin{bmatrix}
		\frac{1}{2} \delta \bbb{\theta} \\
		1
	\end{bmatrix},
\end{equation*}
where $\delta \bbb{\theta} \in \mathbb{R}^3$ represents the angular error. To enhance numerical stability and ensure that the quaternion norm constraint is met, we normalize the quaternion vector after both the propagation step and the update step. In addition to quaternion normalization, we make the covariance matrix symmetric. For more information, the reader is referred to \citep[\S5]{sola2017quaternion}.

\section{Feature Marginalization And State Pruning}\label{section3}
In section \ref{section2}, it was assumed that $m$ number of features are used to perform the update step of the MSCKF algorithm. In this section, the approach used to select these features is explained. In the conventional MSCKF, the update step is carried out when one of the two following cases happen.

\begin{itemize}
	\item \textit{Features are no longer visible.}
	
	This case, which happens most often, occurs when at least one feature is no longer visible in the new image. Each time an image is captured, some of the previously visible features may fall out of the field of view and therefore cannot be tracked in the new image. These features are then used to perform the update step in the algorithm. To maintain the number of features at a constant level, an equivalent number of new features are extracted in the latest image.
	
	Over time, when all features from an image have been utilized in the update steps or are no longer trackable, that image no longer provides any useful information for the algorithm. At this point, the pose associated with that image is removed from the state vector. Additionally, the corresponding row and column related to this pose are pruned from the covariance matrix. This state and covariance pruning is essential as it prevents the state vector and the covariance matrix from growing indefinitely, which would otherwise lead to computational inefficiency and increased memory usage.
	
	\item \textit{State vector reaches a certain size.}
	
	This situation happens when the number of camera poses available in the state vector reaches its maximum, ${N}_{p_{max}}$. This case usually occurs when the camera is not moving fast enough. Consequently, when the camera moves slowly, although some features may fall outside the field of view and be used to carry out an update (previous case), the state vector and the covariance matrix are not pruned. This is due to the fact that the rate at which the features fall outside the field of view is not sufficiently high, so, the sizes of the state vector and the covariance matrix keep increasing. As a result, over time, the number of stored camera poses reaches its maximum, ${N}_{p_{max}}$. In this case, at least one image should be removed. Before removing that image, all its features and their tracks across preceding images are used to perform an update. This ensures that the information contained in those features is not lost. After carrying out the update, the image, along with its corresponding entries in the state vector and the covariance matrix, is removed. Instead of removing only one image, in the original MSCKF, ${N}_{p_{max}}/3$ images are removed. Starting from the second-oldest pose, these ${N}_{p_{max}}/3$ images are evenly spaced in time \citep{ref15}. The reason for keeping the oldest image is that the geometric constraint involving this image usually corresponds to a large baseline, making it carry more valuable information about positioning. According to \citep{ref15}, this method performs well in practice. 
\end{itemize}
Hence, in the conventional MSCKF there is only one tuning parameter which is ${N}_{p_{max}}$.

\begin{figure}[!t]
	\centering
	\includegraphics[width=5.5in]{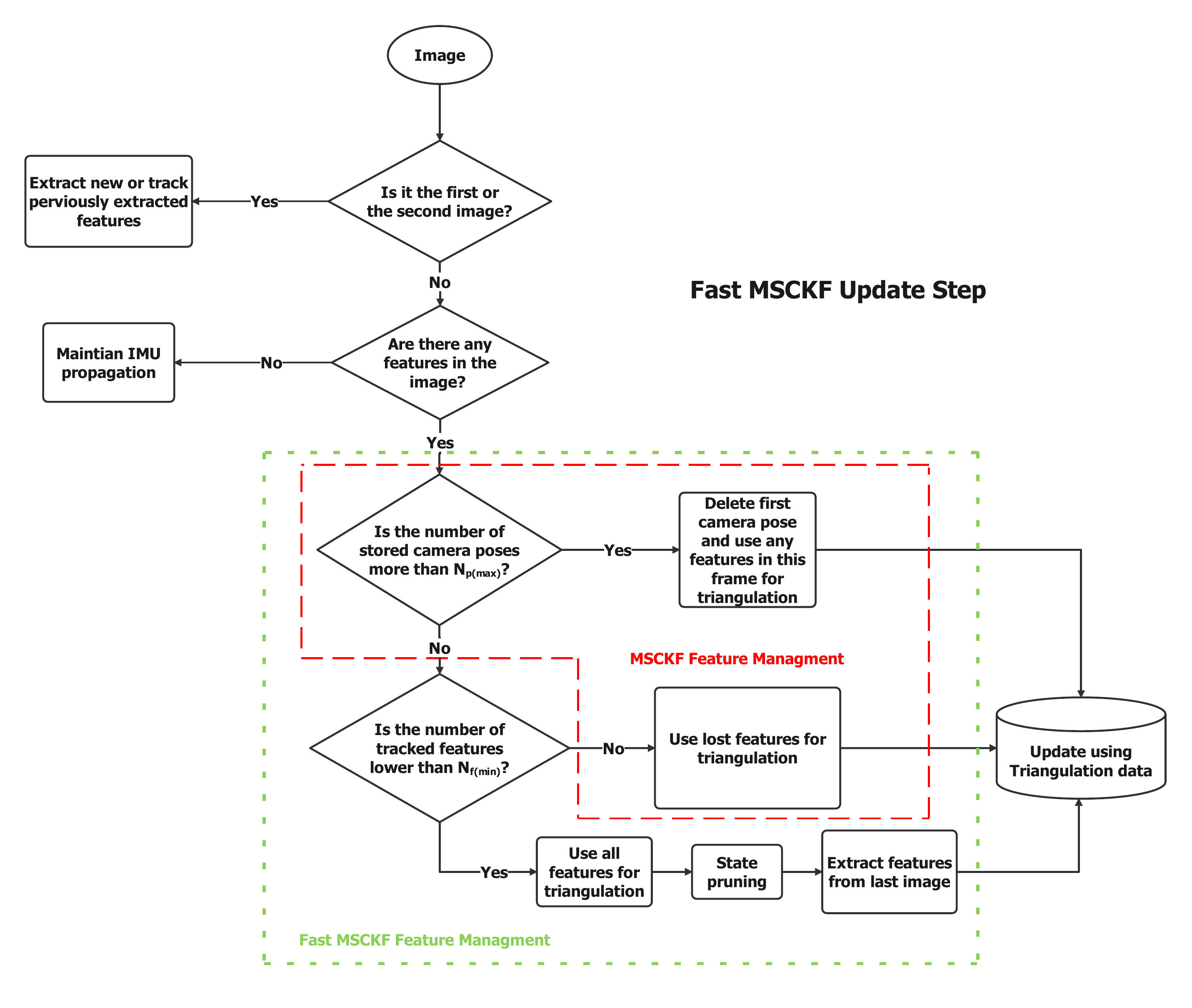}
	\caption{The flowchart of feature management approaches in the MSCKF and the FMSCKF algorithms.}
	\label{fig: flowchart}
\end{figure}

On the other hand, within the Fast MSCKF (FMSCKF) framework, we introduce an additional scenario centered on the minimum number of features that are tracked. In this case, in contrast to the conventional MSCKF approach, where features are extracted from every recorded image, our FMSCKF methodology focuses solely on a subset of images, which we term as \textit{keyframes}.
More specifically, in this approach, once features are extracted from a \textit{keyframe}, they are tracked in the subsequent frames, and no new features are extracted until the number of tracked features falls below a specific threshold, denoted as $N_{f_{min}}$. When the number of tracked features is less than the threshold, all feature tracks are used to perform an update. Post-update, all images, except the last image (the keyframe) are removed, and the state vector and covariance matrix are pruned accordingly. Following that, the final image in the sequence is utilized for the extraction of new features, which will be tracked in forthcoming images.

It is worth mentioning that the concept of keyframes used in this work differs significantly from the common keyframe-based methods found in the literature. Traditionally, keyframe selection is based on the change in the baseline. This means that when the positional change between two consecutive images is large enough, these images are designated as keyframes \citep{ref33}. However, our work adopts a different criterion for defining keyframes. Instead of relying on the change in the baseline, we determine keyframes based on the trackability of features within the frames. Specifically, in our method, a frame $l$ (where $l>1$) is considered a keyframe if the number of features that can be tracked in it falls below a certain threshold, $N_{f_{min}}$. This threshold is a predefined minimum number of features to be tracked, and when the number of these features in a frame drops below this threshold, that frame is marked as a keyframe.

Therefore, within the proposed strategy, three distinct scenarios can trigger an update in the filter, which are as follows in descending order of frequency, from the most frequent to the least frequent.

\begin{itemize}
	\item \textit{Features are no longer visible.}
	
	 The first case is similar to the first case in the conventional MSCKF and occurs when at least one feature exits the field of view of the camera. This case happens most frequently.
	 
	 \item \textit{Feature track reaches a certain number.}
	 
	 The second scenario, which happens second most often, occurs when the number of tracked features falls below a minimum threshold, $N_{f_{min}}$. As explained before, in this case, all feature tracks are used to perform an update. Then, all images in the sequence, except for the most recent one (i.e., the keyframe), are removed. This removal process involves pruning the state vector and the covariance matrix to maintain a manageable size and prevent computational inefficiencies. Following the pruning process, the final image in the sequence, is then used for the extraction of new features. These newly extracted features will be tracked in subsequent images.

	 \item \textit{State vector reaches a certain size.}
	 
	Lastly, the third case, which is equivalent to the second scenario in the conventional MSCKF, happens when the number of camera poses available in the state vector exceeds a maximum number, ${N}_{p_{max}}$. This case happens least often.

\end{itemize}

It is imperative to note that the introduced feature marginalization method has proven to be significantly faster and more accurate in practical applications than the original approach. A schematic of the two feature management methods is shown in Figure \ref{fig: flowchart}. The pseudocode for the update step of the FMSCKF is presented in Algorithm \ref{alg: pseudocode}.

\begin{figure}
	\centering
	\includegraphics[width=5.5in]{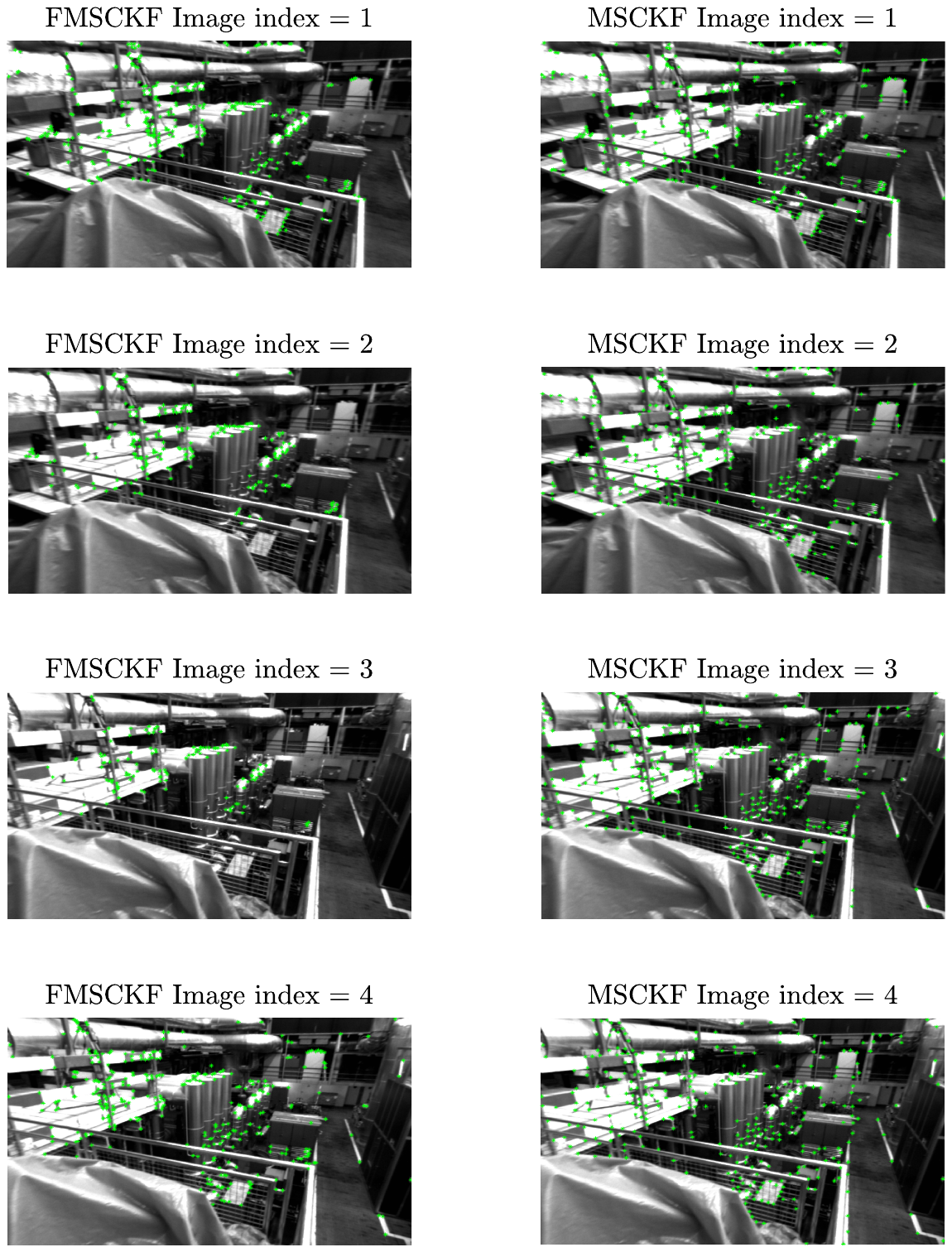}\vspace{-1cm}
	\caption{Comparison of the number of extracted features in the FMSCKF and the MSCKF. In the fourth frame on the left column, the number of tracked features falls below the threshold, and therefore, the algorithm extracts new features. However, the number of features in the original MSCKF algorithm, shown in the right column, is constant. The images are parts of MH\_$01$ dataset \citep{ref27}.}
	\label{fig: features}
\end{figure}
% \end{comment}

To provide a visual comparison between the two feature management policies, Figure \ref{fig: features} shows four consecutive frames and their corresponding features extracted in the MSCKF (right column) and the FMSCKF (left column). As can be seen from Figure \ref{fig: features}, the number of tracked features in the first three frames of the FMSCKF algorithm is decreasing. In the fourth frame, this number falls below the threshold, and as a result, the algorithm extracts new features. On the other hand, the number of features tracked in the MSCKF algorithm remains constant throughout all four frames.
 
% \begin{comment}

\begin{algorithm}
	\caption{Pseudocode for the update step in the proposed algorithm.}\label{alg: pseudocode}
	
	\SetKwData{Left}{left}\SetKwData{This}{this}\SetKwData{Up}{up}
	\SetKwFunction{Union}{Union}\SetKwFunction{FindCompress}{FindCompress}
	\SetKwInOut{Input}{Input}\SetKwInOut{Output}{Output}
	\Input{Augmented state vector, augmented covariance matrix, $m$ features tracked in $n_j$ images.}
	\Output{Updated state vector and covariance matrix.}
	\BlankLine
	\SetAlgoLined
	\uIf{number of stored camera poses $< N_{p_{max}}$}{
		\For{$j = 1, \cdots , m$}{
			\emph{$\bb{Calculate}$ the position of the $j$-th feature in the global frame using triangulation}\;
			
			\For{$i = 1, \cdots, n_j$}{
				\emph{$\bb{Calculate}$ the position of the $j$-th feature in the $i$-th frame using \eqref{eq: estimate feature position}}\;
				\emph{$\bb{Estimate}$ the measurement using \eqref{eq: estimate of z}, then calculate the residual in \eqref{eq: residual}}\;
				\emph{$\bb{Use}$ \eqref{eq: Jacobian wrt feature}, \eqref{eq: Jacobian wrt state}, and \eqref{eq: Jacobian for Hf} to calculate $\prescript{c_i}{}{\bb{H}}_{f_j}$, $\prescript{c_i}{}{\bb{H}}_{X_j}$, and  $\prescript{c_i}{}{\bb{J}}_{f_j}$}\;}
			\emph{$\bb{Use}$ \eqref{eq: complete residual} to \eqref{eq: complete Hx} to compute $\bb{r}_{f_j}$, $\bb{H}_{f_j}$, and $\bb{H}_{X_j}$}\;
			\emph{$\bb{Calculate}$ the left null-space of $\bb{H}_{f_j}$}\;
			\emph{$\bb{Calculate}$ $\bb{H}_{o_j}$, $\bb{r}_{o_j}$, and $\bb{R}_{o_j}$ using \eqref{eq: equation for Hoj} to \eqref{eq: equation for roj}}\;
			\emph{$\bb{Perform}$ a Mahalanobis gating test to reject the outliers}\;}
		\emph{$\bb{Calculate}$ $\bb{H}_o$, $\bb{r}_o$, and $\bb{R}_o$ using \eqref{eq: equation for Ho} to \eqref{eq: equation for Ro}}\;
		\emph{$\bb{Perform}$ a QR-decomposition and calculate $\bb{H}_n$, $\bb{r}_n$, and $\bb{R}_n$, using \eqref{eq: equation for Hn} to \eqref{eq: equation for Rn}}\;
		\emph{$\bb{Update}$ the covariance matrix and the state vector using \eqref{eq: cov matrix update} and \eqref{eq: state update}};   
	}
	\uElseIf{number of tracked features $< N_{\text{fmin}}$}{
		\emph{$\bb{For}$ all features run 2 - 16}\;
		\emph{$\bb{Prune}$ the state vector and the covariance matrix}\;
		\emph{$\bb{Extract}$ new features from the last image}\;
	}
	\Else{
		\emph{$\bb{Delete}$ the first camera pose from the state vector, and its corresponding entries from the covariance matrix}\;
		\emph{$\bb{For}$ all features in the deleted frame run 2 - 16};\
	}
	
\end{algorithm}

\begin{table}[h]
	\caption{The update rate (in $Hz$) for various algorithms implemented on the EuRoC MAV dataset \citep{ref27}. The codes were executed on MATLAB.} \label{table1}
	\begin{tabular}{|c|c|c|c|c|c|c|}
		\hline
		\rule{0pt}{4ex}
		Dataset & ROVIO & SVO (VIO) & SVO+GTSAM & MSCKF  & VINS\_Mono &  FMSCKF \\ \hline
		\rule{0pt}{3ex}
		MH\_$01$        & $38.75$   & $32.51$    &$64.67$           & $28.49$     &$35.31$ & $\bb{106.38}$\\ 
		\hline
		\rule{0pt}{3ex}
		MH\_$02$        & $41.98$    & $35.13$    & $70.06$            & $30.86$      &$36.12$  &$ \bb{117.65}$\\
		 \hline
		\rule{0pt}{3ex}
		MH\_$03$        & $37.36$    & $31.45$    & $62.36$           & $27.47$    &$35.06$ & $\bb{93.46}$\\
		 \hline
		\rule{0pt}{3ex}
		MH\_$04$        & $44.01$     & $37.1$   & $73.46$            & $32.36$     &$35.31$ & $\bb{101.01} $\\
		 \hline
		\rule{0pt}{3ex}
		MH\_$05 $       & $40.72$    & $34.35$    & $67.96$            &$29.94$    &$35.82$ &$ \bb{114.94} $\\
		 \hline
		\rule{0pt}{3ex}
		VR1\_$01$       & $34.43$     & $29.23$   & $57.47$            & $25.32$     &$32.87$ & $\bb{84.75}$\\ 
		\hline
		\rule{0pt}{3ex}
		VR1\_$02$       & $39.08$    & $32.91$    & $65.23$            & $28.74$     &$35.46$  & $\bb{107.53} $\\ 
		\hline
		\rule{0pt}{3ex}
		VR1\_$03$       & $37.06$     & $31.12$   & $61.85$            & $27.25$     &$34.73$ & $\bb{99.01}$\\ 
		\hline
		\rule{0pt}{3ex}
		VR2\_$01 $      & $38.64$    & $33.19$    & $64.49$            &$28.41$    &$35.73$ & $\bb{111.11}$ \\ 
		\hline
		\rule{0pt}{3ex}
		VR2\_$02 $      &$34.78$    & $30.54$    & $58.06$           & $25.58$    &$35.04$ &$ \bb{96.15}$\\
		 \hline
		\rule{0pt}{3ex}
		VR2\_$03$       & $36.66$    & $31.43$    & $61.19$           & $26.95$     &$35.17$ &$ \bb{103.09}$\\
		 \hline
	\end{tabular}
\end{table}

\section{Results}\label{section4}
In this section, the results of the proposed algorithm are presented. For the original MSCKF \citep{ref15}, the maximum allowable number of poses is chosen to be ${N}_{p_{max}} = 20$. For the FMSCKF, the minimum number of features is set to $N_{f_{min}} = 8$, and similar to the original MSCKF, the maximum allowable number of poses is chosen to be ${N}_{p_{max}} = 20$. The rationale for selecting $N_{f_{min}} = 8$ is elaborated in Section \ref{section5}. In the keyframes, a maximum of $350$ Harris corners are extracted and tracked in consecutive frames. To track the extracted features, the well-known KLT algorithm \citep{ref28} is used. The Random Sample Consensus (RANSAC) algorithm is used to reject outliers in feature tracking \citep{ref29}.

\subsection{Public Dataset}
The EuRoC MAV dataset \citep{ref27} is used to evaluate the algorithm. To provide a comprehensive comparison of the update rate, the results of SVO (VIO) \citep{ref30}, SVO+GTSAM \citep{ref6}, VINS-mono \citep{ref9} and ROVIO \citep{ref12} algorithms are presented. Default settings are applied for these algorithms. All computations were conducted on a personal MacBook Air equipped with an M1 chip and 8GB Memory.

\begin{figure}[!t]
	\centering
	\includegraphics[width=3.5in]{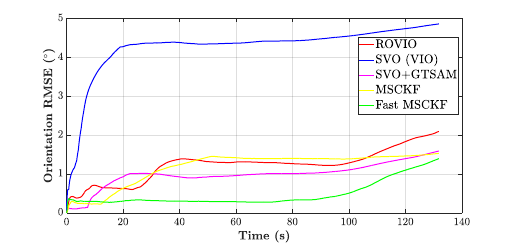}
	\caption{orientation estimation RMSE for different algorithms.}
	\label{fig: ori rmse}
\end{figure}

\begin{figure}[!t]
	\centering
	\includegraphics[width=3.5in]{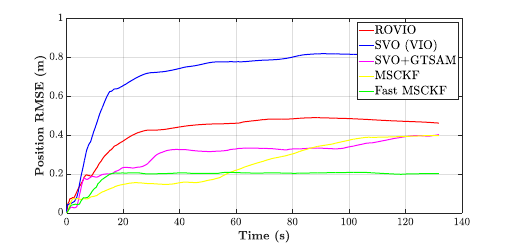}
	\caption{position estimation RMSE for different algorithms.}
	\label{fig: pos rmse}
\end{figure}

\begin{figure}[!t]
	\centering
	\includegraphics[width=5in]{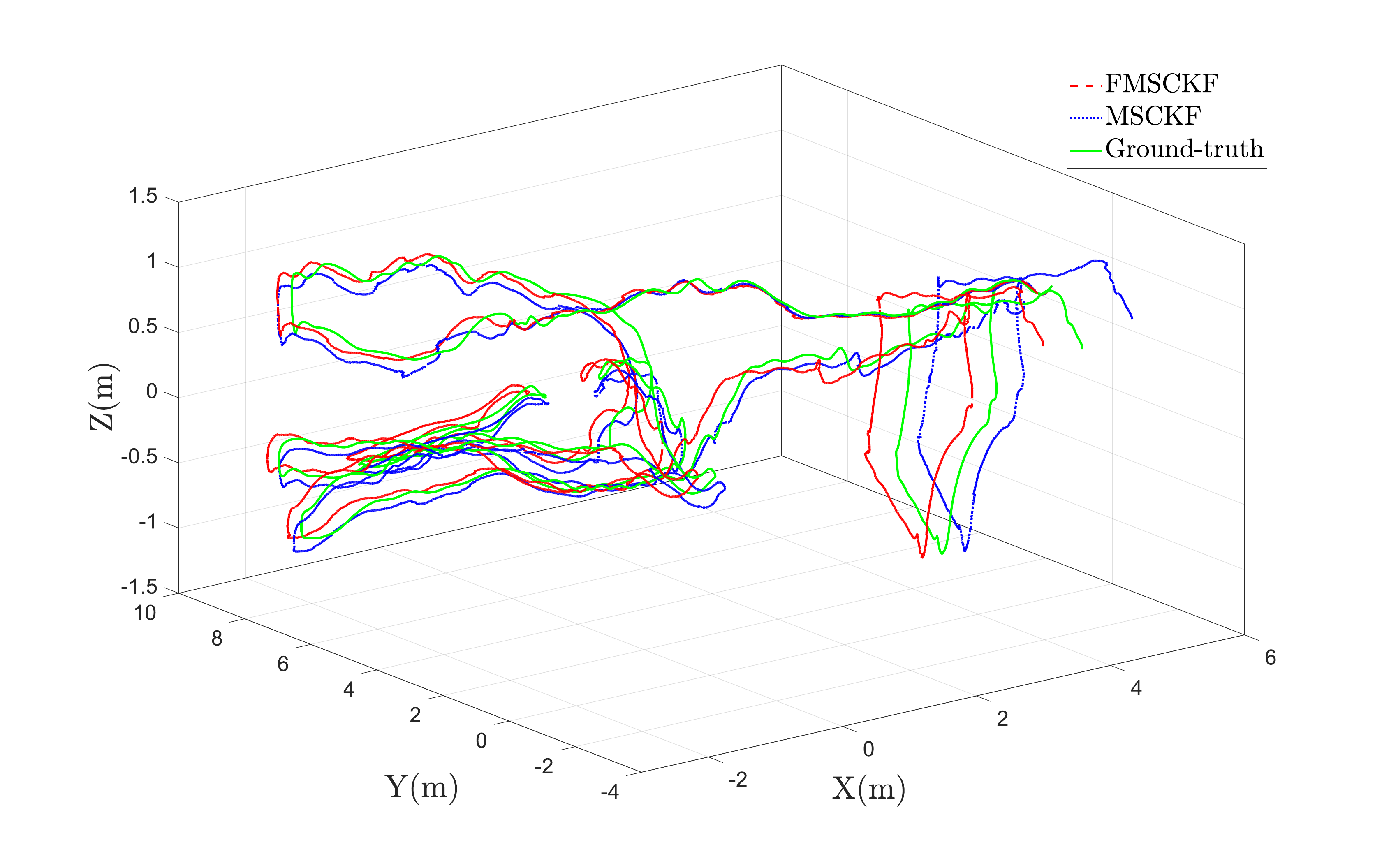}
	\caption{Estimated 3D paths for MH\_$01$ using the MSCKF (blue) and the FMSCKF (red). The ground-truth path is plotted in green \citep{ref27}.}
	\label{fig: 3d path}
\end{figure}

The update rates for different algorithms are presented in Table \ref{table1}. The superior performance of the FMSCKF compared to other algorithms in terms of update rate is evident from Table \ref{table1}. To visually compare the algorithms, the Root Mean Square Error (RMSE) plots in position and orientation estimation for one dataset (MH\_$01$) are shown in Figures \ref{fig: ori rmse} and \ref{fig: pos rmse}, respectively. From Figure \ref{fig: ori rmse}, it can be seen that the FMSCKF outperforms all the other algorithms in terms of orientation estimation. One should note that the ascending behavior of orientation error in Figure \ref{fig: ori rmse} is due to the fact that in EKF-based visual-inertial estimators, the yaw angle is not observable \citep{ref31} and therefore, it drifts over time. From Figure \ref{fig: pos rmse}, it can be deduced that the FMSCKF has a lower error in position estimation compared to other algorithms. Additionally, in Figure \ref{fig: 3d path}, the 3D paths for one dataset (MH\_$01$) using the MSCKF and the FMSCKF, alongside the ground-truth are shown. Final errors in orientation and position estimation for these algorithms are provided in Table \ref{table2}.

\begin{figure}[!t]
	\centering
	\includegraphics[width=4.5in]{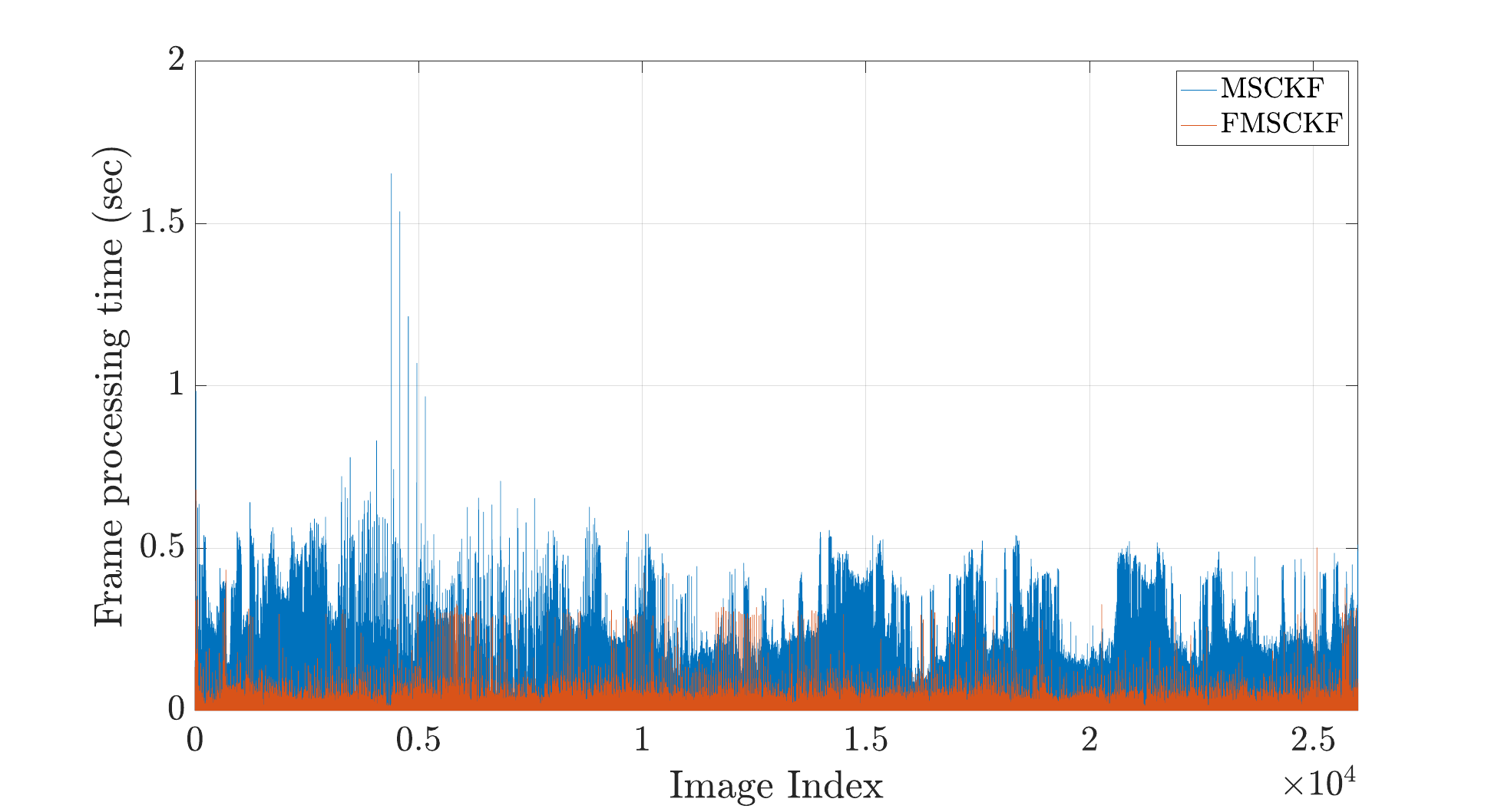}
	\caption{Frame processing time for the FMSCKF (red) and the MSCKF (blue).}
	\label{fig: process time}
\end{figure}

Frame processing time for MH\_$01$ is shown in Figure \ref{fig: process time}. As expected, the FMSCKF processes the frames significantly faster than the original MSCKF.

\begin{figure}
	\centering
	\includegraphics[width=6.5in]{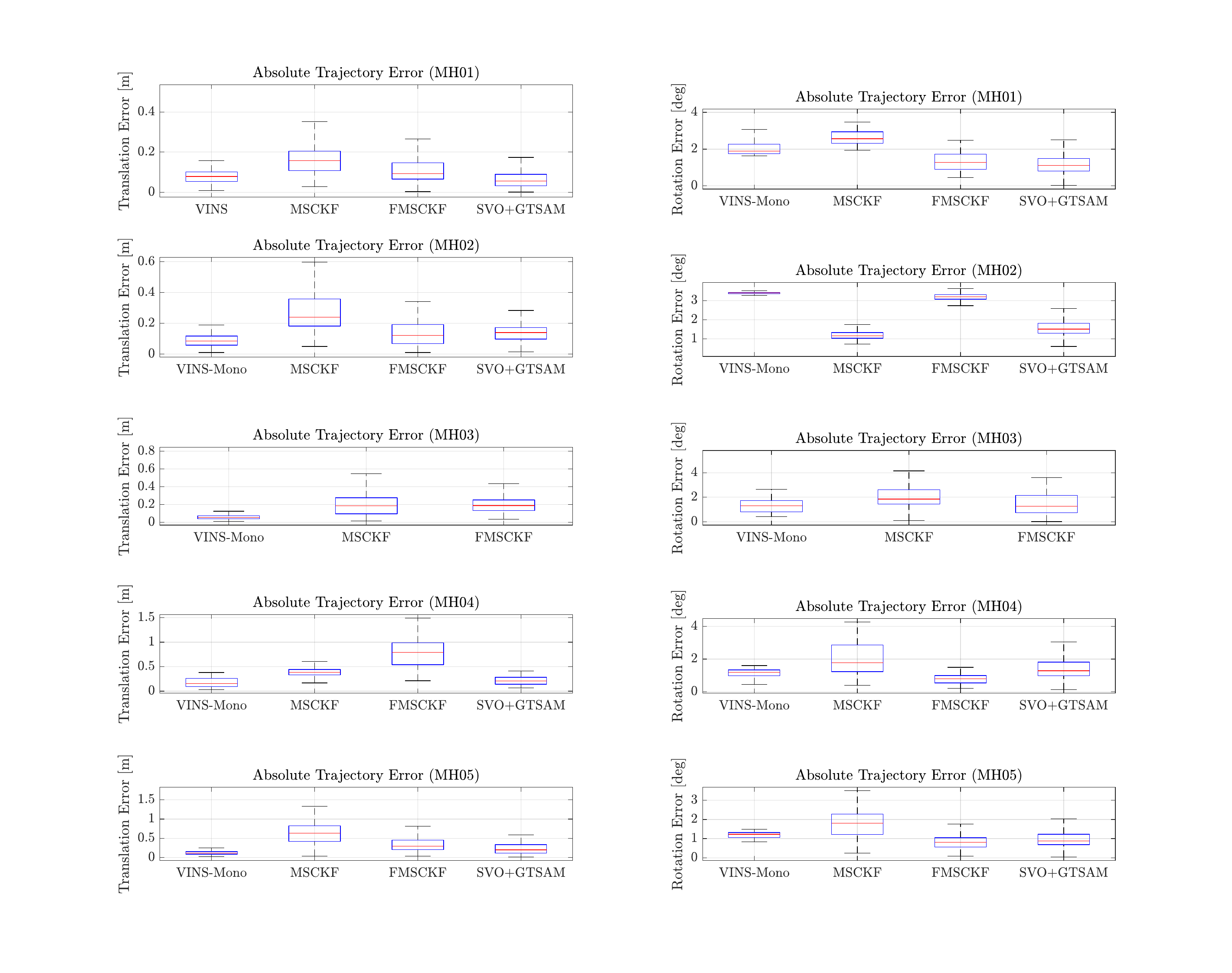}
	\caption{Absolute Trajectory Error (ATE) for various algorithms.}
	\label{fig: ate}
\end{figure}

\begin{figure}
	\centering
	\includegraphics[width=4.5in]{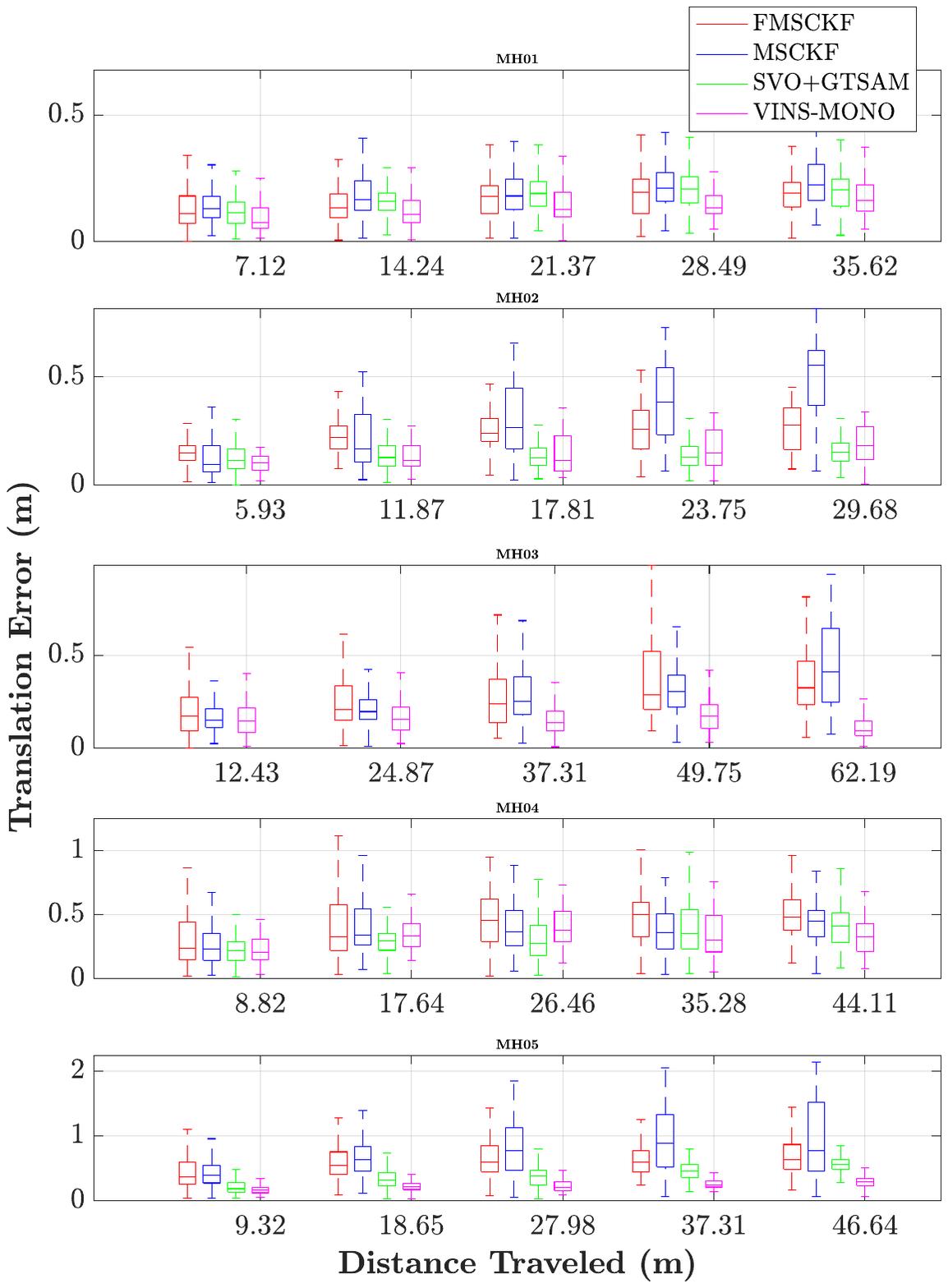}
	\caption{Relative Trajectory  Error (RTE) for various algorithms.}
	\label{fig: rte}
\end{figure}

Moreover, box plots illustrating Absolute Trajectory Error (ATE) and Relative Trajectory Error (RTE) are depicted in Figures \ref{fig: ate} and \ref{fig: rte}, respectively. As can be seen in Figures \ref{fig: ate} and \ref{fig: rte}, the FMSCKF demonstrates accuracy on par with other algorithms. Notably, it frequently surpasses the MSCKF in both ATE and RTE, while concurrently achieving a significantly higher output rate, as shown in Table \ref{table1}. Note that we did not include the ATE and RTE results of SVO+GTSAM for MH\_$03$ in Figures \ref{fig: ate} and \ref{fig: rte}, as its errors were notably higher than compared to the other algorithms.

\begin{figure}
	\centering
	\includegraphics[width=4.5in]{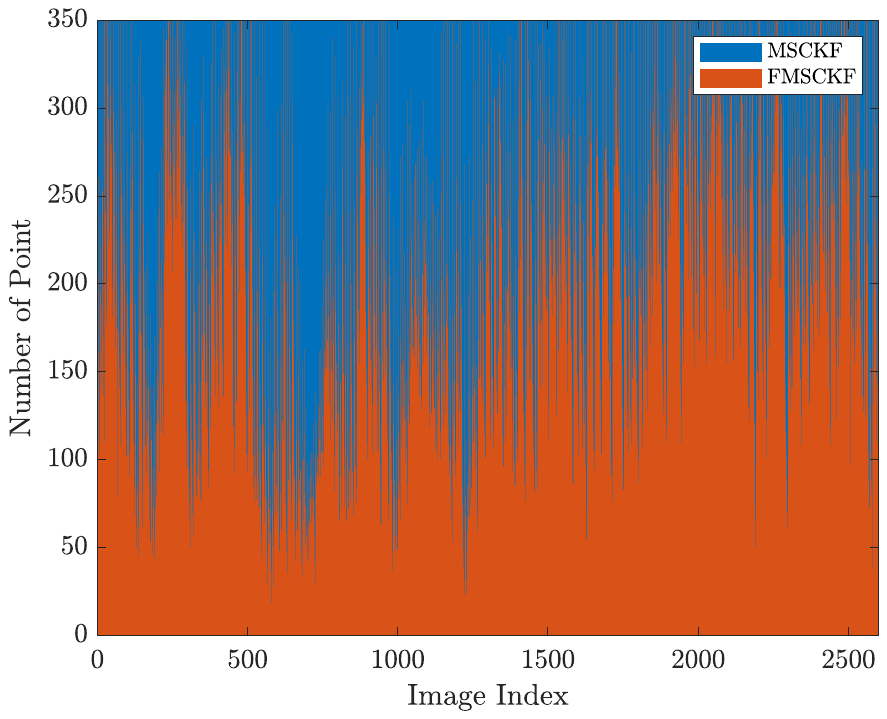}
	\vspace{-3cm}
	\caption{Number of tracked features in each image for the MSCKF (blue) and the FMSCKF (red).}
	\label{fig: features tracked}
\end{figure}

\begin{figure}
	\centering
	\includegraphics[width=4.5in]{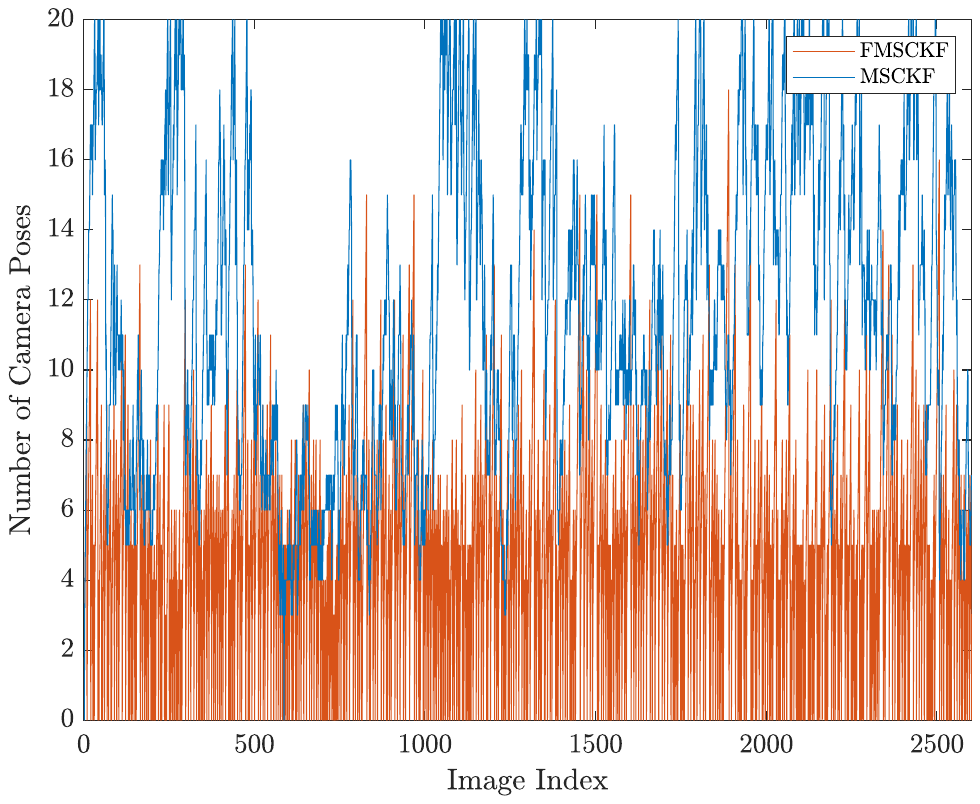}
	\vspace{-4cm}
	\caption{Number of camera poses stored in the state vector for the FMSCKF (blue) and the MSCKF (red).}
	\label{fig: num pos}
\end{figure}

\begin{figure}
	\centering
	\includegraphics[width=4.5in]{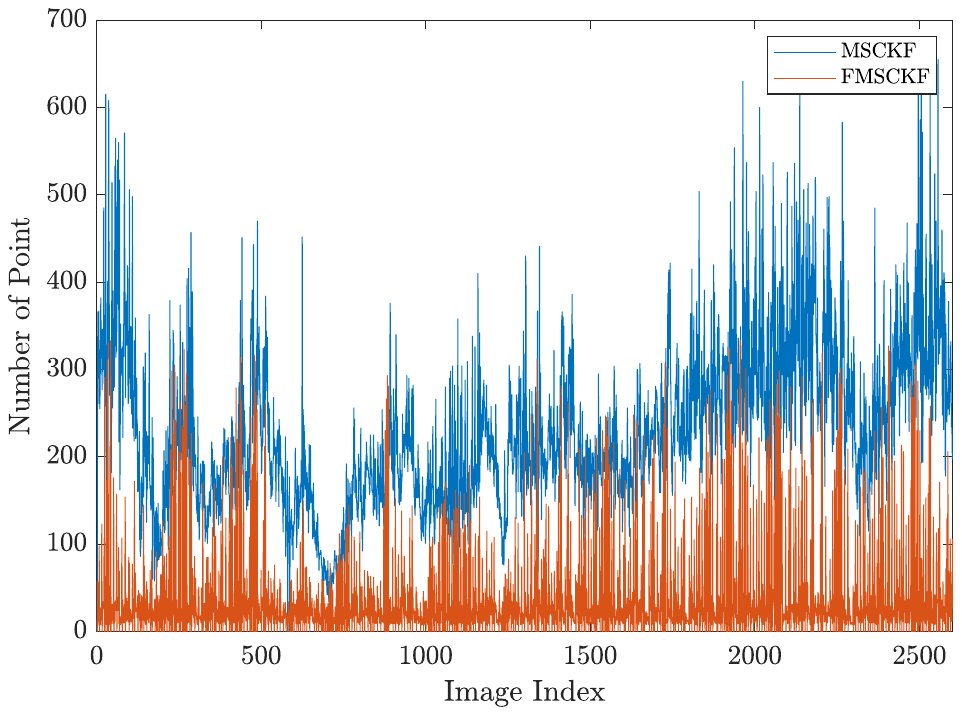}
	\vspace{-4cm}
	\caption{Number of features used in the update step of the filter the MSCKF (blue) and the FMSCKF (red).}
	\label{fig: features used}
\end{figure}

Furthermore, the number of tracked features, camera poses, and features used for update are shown in Figures \ref{fig: features tracked}, \ref{fig: num pos}, and \ref{fig: features used}, respectively. As evident, both the number of tracked features and the number of features used for update in the FMSCKF are lower than those in the MSCKF, which, as claimed, results in a higher update rate of the FMSCKF.

\begin{table}[h]
	\caption{Results of various algorithms for MH\_$01$ \citep{ref27}. The final point error is calculated with respect to the traveled distance.\label{table2}}
	\centering
	\begin{tabular}{|c|c|c|}
		\hline
		\rule{0pt}{4ex}
		Method & \parbox{26mm}{Final point error ($\%$)} & \parbox{30mm}{Final orientation error $\left(o\right)$}\\
		\hline
		\rule{0pt}{3ex}
		ROVIO &	$0.57$ & $2.09$ \\
		\hline
		\rule{0pt}{3ex}
		SVO (VIO) & $0.98$ & $4.86$\\
		\hline
		\rule{0pt}{3ex}
		SVO+GTSAM & $0.49$ & $1.59$\\
		\hline
		\rule{0pt}{3ex}
		MSCKF & $0.49$ & $1.52$\\
		\hline
		\rule{0pt}{3ex}
		FMSCKF &$\bb{0.25}$ & $\bb{1.39}$\\
		\hline
	\end{tabular}
\end{table}

\subsection{Real-world Experiments}
\begin{figure}[!t]
	\centering
	\includegraphics[width=3.5in]{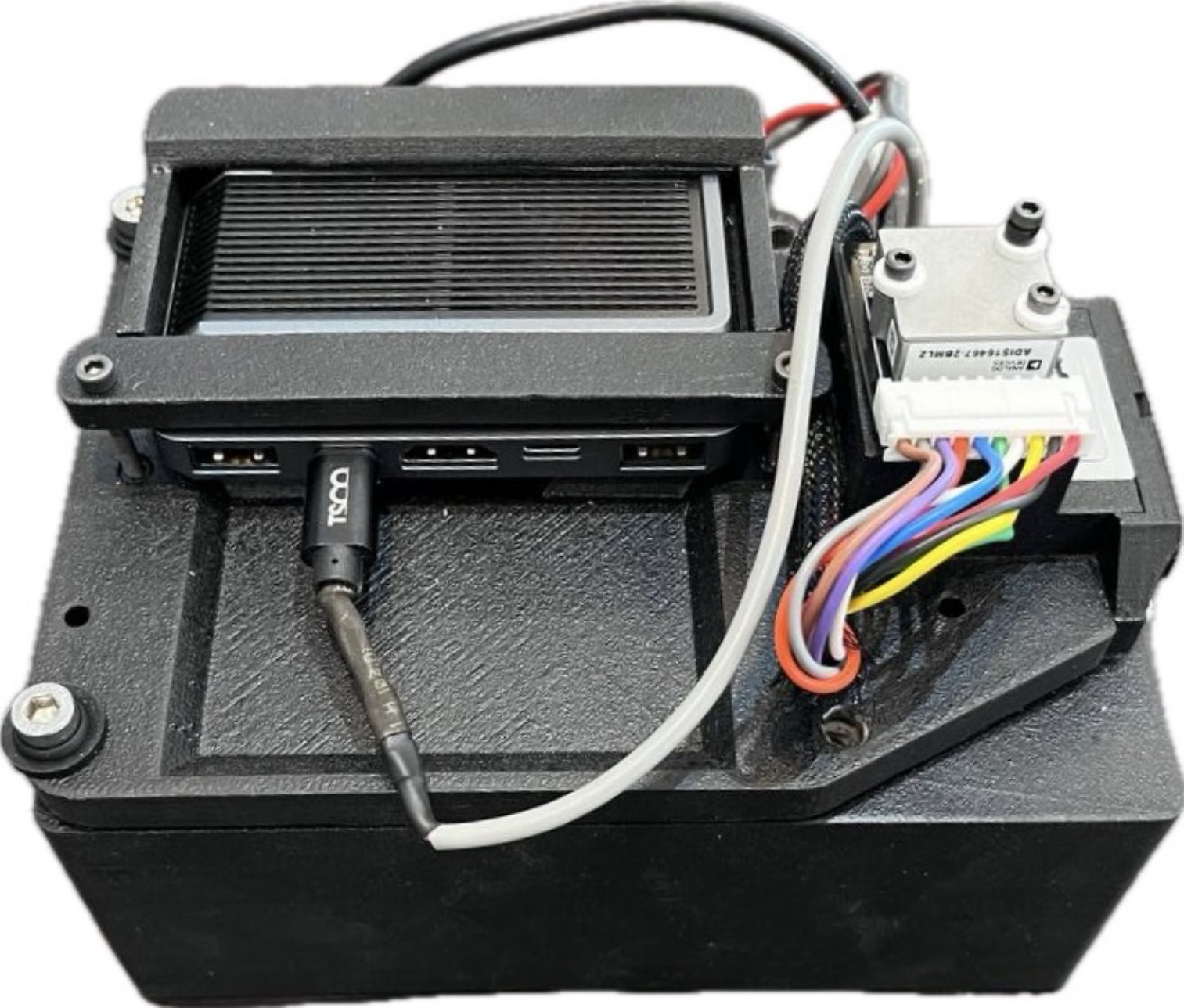}
	\caption{The customized sensor setup. The camera is mounted on the front side. An STM32 F103c8t6 receives data from the IMU, and after an initial processing, sends it to the Nvidia Jetson board.}
	\label{fig: setup}
\end{figure}

An experimental setup was developed and utilized to test the proposed algorithm. The main processor utilized was an NVIDIA Jetson Xavier NX. The setup is shown in Figure \ref{fig: setup}. Further details regarding the sensors employed in the platform can be found in Table 1 \ref{table4}. The setup was calibrated using Kalibr library \citep{ref32}. The algorithms were not executed in real-time; instead, the acquired data was later used on the personal laptop to run the filters. Three experiments were conducted, the results of which will be presented in the following subsections. 

\begin{table}
	\caption{The sensor setup used in this paper.}
	\centering
	\label{table4}
	\begin{tabular}{|c|c|c|}
		\hline
		\rule{0pt}{4ex}
		Sensor & Type & Output rate ($Hz$)\\
		\hline
		\rule{0pt}{4ex}
		ADIS $16467-2$ & MEMS IMU & $100$\\
		\hline
		\rule{0pt}{4ex}
		Basler acA$1300-200$uc & Global shutter camera & $10$\\
		\hline
	\end{tabular}
\end{table}

\subsubsection{Short-range experiment}\label{sub4-2-1}
\begin{figure}[!t]
	\centering
	\includegraphics[width=3in]{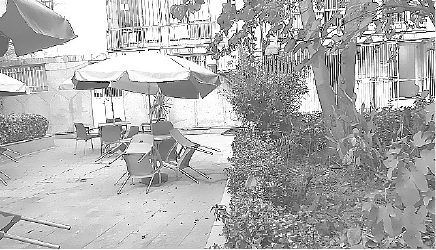}
	\caption{A sample image of the short-range dataset.}
	\label{fig: sample short}
\end{figure}

In the first experiment, a path spanning 45 meters was traversed in approximately 90 seconds. A sample image recorded during this experiment is shown in Figure \ref{fig: sample short}. For a more precise and explicit comparison, only the original MSCKF and the proposed FMSCKF were employed for this experiment.

\begin{figure}[!t]
	\centering
	\includegraphics[width=3.5in]{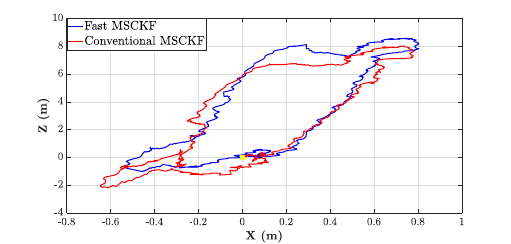}
	\caption{Estimated path in XZ plane using the FMSCKF (blue) and the MSCKF (red) in the short-range experiment. The starting point is indicated by a yellow square and final points are marked with circles.}
	\label{fig: short xz}
\end{figure}

The setup was moved by a person along a random path, making it impossible to recover the ground-truth trajectory. As a result, to evaluate the performance of the two algorithms, we attempted to create a loop-shaped path and return precisely to the starting point. This enabled us to measure and compare the final point error for each algorithm. The estimated trajectories are shown in Figure \ref{fig: short xz}. As evident from Figure \ref{fig: short xz}, both algorithms exhibit good performance in position estimation. It is worth mentioning that the difference in the estimated trajectories of the two algorithms is primarily due to the low quality of the camera.

The final position estimates of the MSCKF and the FMSCKF, as well as the ground truth, are listed in Table \ref{table5}.

\begin{table}
	\caption{Results of the short-range experiment.}
	\centering
	\label{table5}
	\begin{tabular}{|c|c|c|c|c|}
		\hline
		\rule{0pt}{4ex}
		Algorithm & $x_f (m)$ & $y_f (m)$ & $z_f (m)$ & Final point error ($\%$)\\
		\hline
		\rule{0pt}{4ex}
		FMSCKF & $0.0806$ & $0.0355$ & $0.1085$ & $0.31$\\
		\hline
		\rule{0pt}{4ex}
		MSCKF & $0.0899$ & $-0.1385$ & $0.1747$ & $0.53$\\
		\hline
		\hline
		Ground truth & $0$ & $0$ & $0$ & N.A.\\
		\hline
	\end{tabular}
\end{table}

\begin{figure}[!t]
	\centering
	\includegraphics[width=3.5in]{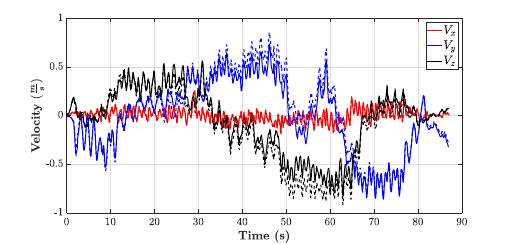}
	\caption{Estimated velocities during the short-range experiment using the FMSCKF (solid) and the MSCKF (dashed).}
	\label{fig: short vel}
\end{figure}

\begin{figure}[!t]
	\centering
	\includegraphics[width=3.5in]{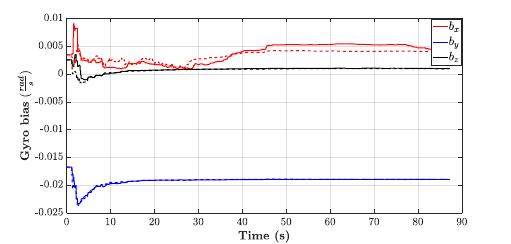}
	\caption{Estimated gyroscope biases during the short-range  experiment using the FMSCKF (solid) and the MSCKF (dashed).}
	\label{fig: short gyro bias}
\end{figure}

\begin{figure}[!t]
	\centering
	\includegraphics[width=3.5in]{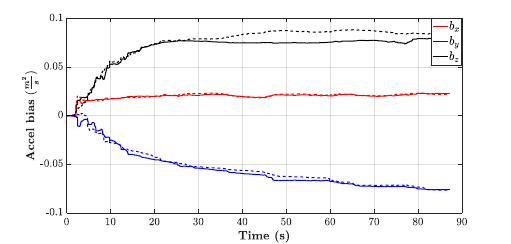}
	\caption{Estimated accelerometer biases during the short-range  experiment using the FMSCKF (solid) and the MSCKF (dashed).}
	\label{fig: short accel bias}
\end{figure}

Accordingly, the final point error of the FMSCKF is $0.14(m)$, while that of the MSCKF is $0.24(m)$. In other words, for the 45-meter-long trajectory, the final point error of the FMSCKF is $0.31\%$ and the final point error of the MSCKF is $0.53\%$ of the traveled distance. These errors are in good compliance with the final point errors of the algorithms running on the EuRoC MAV dataset. For instance, for MH\_$01$, these numbers are $0.25\%$ and $0.49\%$, respectively. Estimated velocities and biases are plotted in Figures \ref{fig: short vel} to \ref{fig: short accel bias}.

\subsubsection{Mid-range experiment}\label{sub4-2-2}
\begin{figure}[!t]
	\centering
	\includegraphics[width=3in]{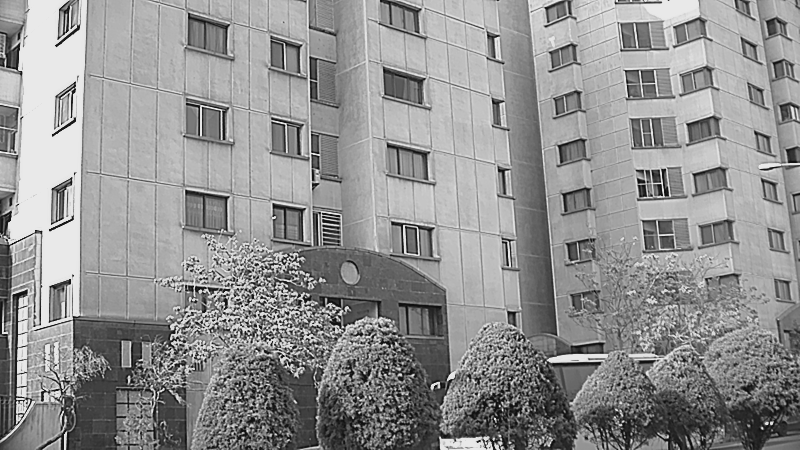}
	\caption{A sample image of the mid-range dataset.}
	\label{fig: mid sample}
\end{figure}

\begin{figure}[!t]
	\centering
	\includegraphics[width=4in]{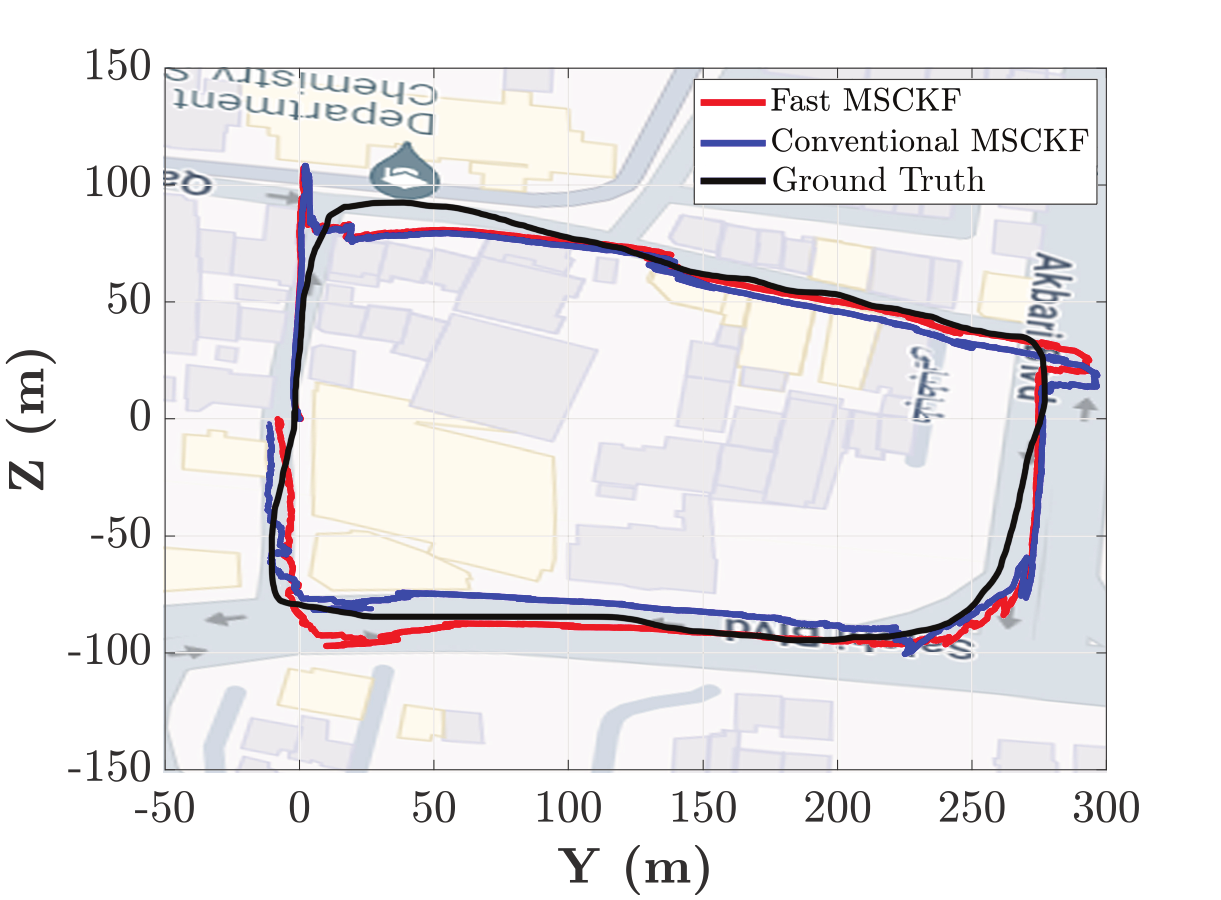}
	\caption{Estimated path in YZ plane using the MSCKF (blue) and the FMSCKF (red) along with the ground-truth path (black) in the mid-range experiment.}
	\label{fig: mid yz}
\end{figure}

In the second experiment, the setup was mounted on a car, which then traversed a 900-meter-long path. A sample image recorded during this experiment is shown in Figure \ref{fig: mid sample}. The estimated paths using the FMSCKF and the MSCKF along with the ground-truth path are shown in Figure \ref{fig: mid yz}. Note that, similar to the short-range test, GPS signal was inaccurate due to presence of tall buildings. However, unlike the short-range test, the ground-truth was obtained by reconstructing the trajectory of the vehicle based on the lanes it traveled in. It is worth noting that since this ground-truth reconstruction is prone to various errors, it shall not be used for numerical evaluation of the algorithms. It is provided solely for visual comparison purposes. The final position estimates of the MSCKF and the FMSCKF, along with the ground truth, are listed in Table \ref{table6}.

\begin{table}
	\caption{Results of the mid-range experiment.}
	\centering
	\label{table6}
	\begin{tabular}{|c|c|c|c|c|}
		\hline
		\rule{0pt}{4ex}
		Algorithm & $x_f (m)$ & $y_f (m)$ & $z_f (m)$ & Final point error ($\%$)\\
		\hline
		\rule{0pt}{4ex}
		FMSCKF & $0$ & $-3.3210$ & $-0.7822$ & $0.38$\\
		\hline
		\rule{0pt}{4ex}
		MSCKF & $0$ & $-4.512$ & $-1.128$ & $0.51$\\
		\hline
		\hline
		Ground truth & $0$ & $0$ & $0$ & N.A.\\
		\hline
	\end{tabular}
\end{table}

\subsubsection{Long-range experiment}\label{sub4-2-3}
\begin{figure}[!t]
	\centering
	\includegraphics[width=3in]{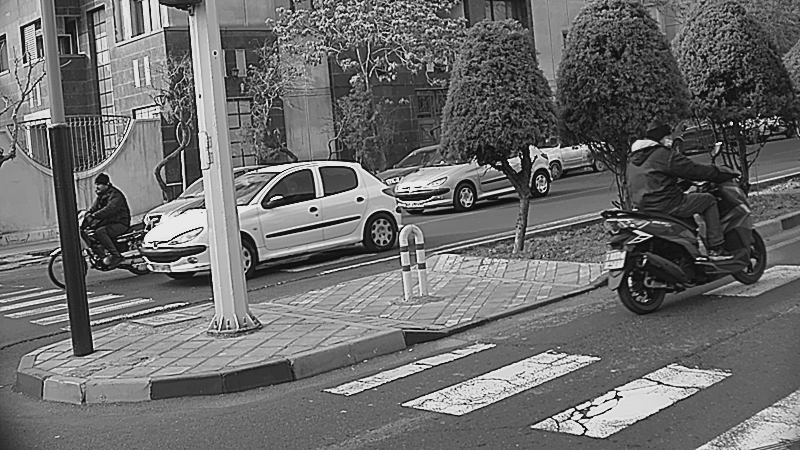}
	\caption{A sample image of the long-range dataset.}
	\label{fig: long sample}
\end{figure}

\begin{figure}[!t]
	\centering
	\includegraphics[width=4in]{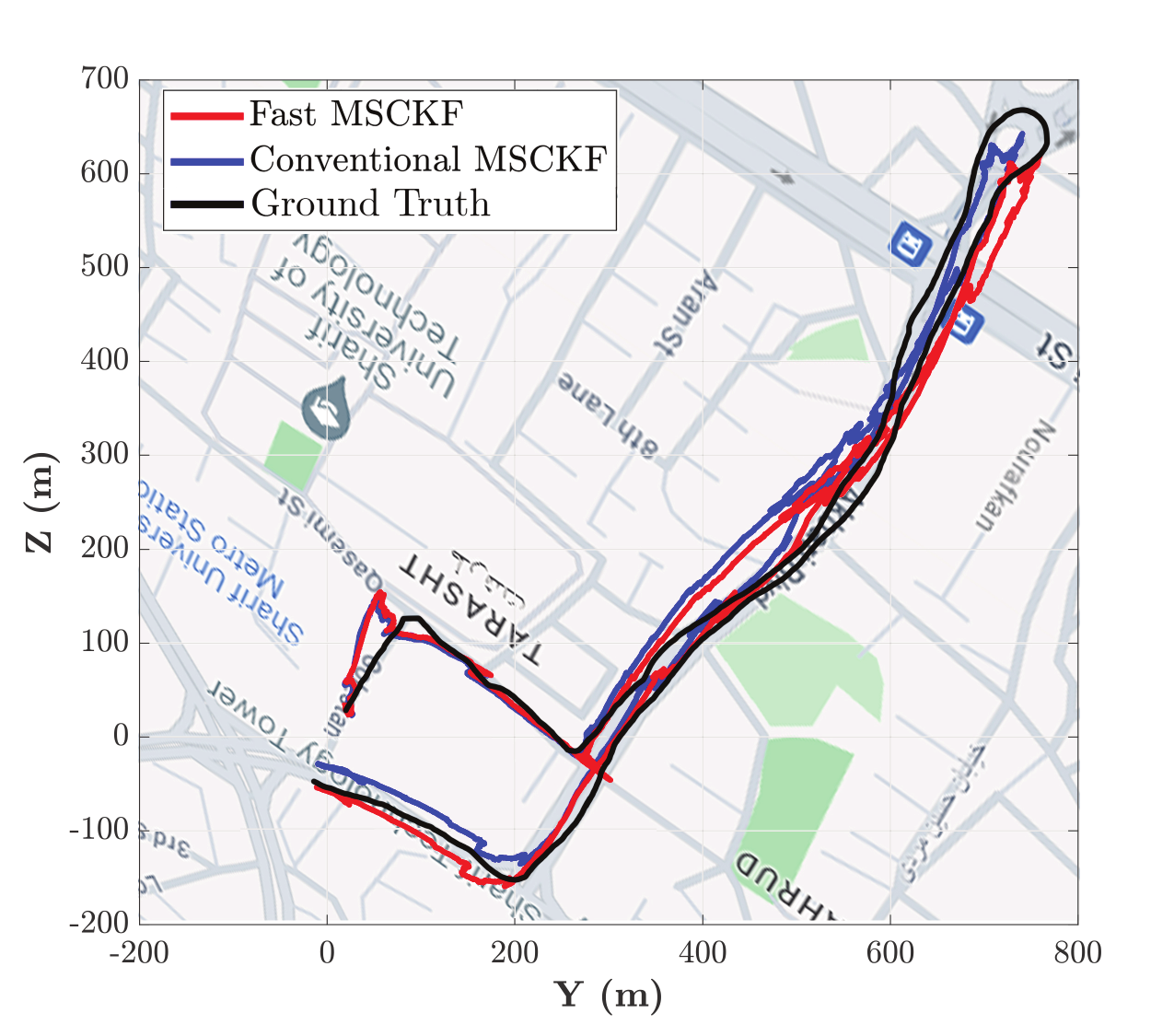}
	\caption{Estimated path in YZ plane using the FMSCKF (red) and the MSCKF (blue) along with the ground-truth path (black) in the long-range experiment.}
	\label{fig: long yz}
\end{figure}

In the last experiment, we traversed a path spanning 2500 meters in a crowded neighborhood with a large number of moving cars and people. A sample image recorded during this experiment is shown in Figure \ref{fig: long sample}. The ground-truth path along with the estimated paths using both the FMSCKF and the MSCKF algorithms is shown in Figure \ref{fig: long yz}. Similar to the mid-range experiment, the ground-truth was obtained by reconstructing the trajectory of the vehicle based on the lanes it traveled in.

The final position estimates of the MSCKF and the FMSCKF, as well as the ground truth, are listed in Table \ref{table7}. Using the final point estimates in Table \ref{table7}, it is straightforward to calculate the final point error for the algorithms. For the FMSCKF it is $0.41\%$, and for the MSCKF, it is $1.02\%$ of the traveled distance. We observe that both algorithms perform well in long distances and in the presence of dynamic objects.

\begin{table}
	\caption{Results of the long-range experiment.}
	\centering
	\label{table7}
	\begin{tabular}{|c|c|c|c|c|}
		\hline
		\rule{0pt}{4ex}
		Algorithm & $x_f (m)$ & $y_f (m)$ & $z_f (m)$ & Final point error ($\%$)\\
		\hline
		\rule{0pt}{4ex}
		FMSCKF & $0$ & $-28.5850$ & $-55.5651$ & $0.41$\\
		\hline
		\rule{0pt}{4ex}
		MSCKF & $0$ & $-30.3751$ & $-27.3251$ & $0.51$\\
		\hline
		\hline
		Ground truth & $0$ & $-20$ & $-50$ & N.A.\\
		\hline
	\end{tabular}
\end{table}

\begin{comment}
\begin{table}[h]
		\caption{Experimental test of the FMSCKF in real-world scenarios}
	\begin{tabular}{|c|c|c|c|c|c|}
		\hline
		\rule{0pt}{4ex}
		\parbox{25mm}{Range}             &\parbox{8mm}{Final x(m)}  &  \parbox{8mm}{Final y(m)} & \parbox{8mm}{Final z(m)}  & \parbox{20mm}{Final point error (\%)} &\parbox{20mm}{Update Rate (Hz)}  \\ \hline
		\rule{0pt}{4ex}
		\parbox{25mm}{Short-Range ($45m$)} & $0.0806$     & $0.0355$     & $0.1085$     &$ 0.31$                   & $98.22 $           \\ \hline
		\rule{0pt}{4ex}
		\parbox{25mm}{Mid-Range ($900m$)-$2D$ } &   $0 $  & $-0.7822$    & $-3.321$          & $0.38$                   & $101.25$           \\ \hline
		\rule{0pt}{4ex}
		\parbox{25mm}{Long-Range ($2500m$)-$2D$} &  $0$    & $-8.585$     & $5.565 $          & $0.41$                   & $99.21$            \\ \hline
	\end{tabular}
	\label{tab:my-table2}
\end{table}
\end{comment}

\section{Discussion}\label{section5}
The results in section \ref{section4} demonstrate the superiority of the FMSCKF over the original MSCKF in terms of both speed and accuracy. This superiority is explained in this section. The enhanced update rate in the FMSCKF can be attributed to two key factors. Firstly, the second scenario in the FMSCKF, where the number of features reaches its minimum, happens more frequently than the third one. Consequently, our method conducts more frequent pruning of the state vector and the covariance matrix compared to the original MSCKF, leading to a noticeable enhancement in algorithm speed. Secondly, in the conventional MSCKF, features are extracted from all frames, but in the FMSCKF, features are only extracted from the keyframes, when the number of tracked features is less than $N_{f_{min}}$. This strategic feature extraction methodology significantly reduces the computational cost associated with image processing in the FMSCKF.

As illustrated in Section \ref{section4}, our approach exhibits comparable accuracy to the MSCKF in orientation estimation while surpassing it in position estimation. The reason for the higher accuracy in position in the FMSCKF is twofold. Firstly, the rate at which updates occur in the FMSCKF is higher compared to the original MSCKF. To elaborate, as mentioned in section \ref{section2b}, in the augmentation step, camera poses are calculated using the propagated IMU states. We know that the propagated states drift quickly, as IMU readings are integrated in that step. As a result, the longer the filter delays the update step, the less accurate camera poses will become, and consequently, the less accurate the triangulated points will be. Secondly, in the FMSCKF, when the number of tracked features falls below $N_{f_{min}}$, a relatively large number of features $\left(N_{f_{min}}\right)$ are used to do the update. Hence, the updated states are more accurate.  It is worth noting that this particular event does not occur within the original MSCKF methodology.

\begin{figure}[h]
	\centering
	\includegraphics[width=3.5in]{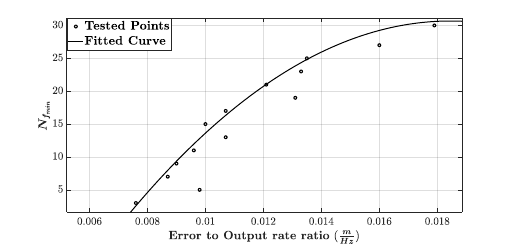}
	\caption{The effect of $N_{f_{min}}$on the error to speed ratio in the FMSCKF.}
	\label{fig: nfmin}
\end{figure}

As discussed earlier, $N_{f_{min}}$ is a tuning parameter and must be set appropriately. Setting $N_{f_{min}}$to high values causes high state pruning rates, thereby reducing the algebraic computational cost. However, a high value of $N_{f_{min}}$ increases feature extraction frequency, which leads to a significantly higher image processing cost. Our extensive tests showed that the algebraic cost constitutes only $10\%$ of the total computational cost, and the remaining $90\%$ pertains to the image processing part. Moreover, a lower value of $N_{f_{min}}$ corresponds to increased accuracy. This phenomenon occurs because when the minimum number of features is limited, the algorithm gains more information about individual features over consecutive image frames. In essence, the algorithm's knowledge about a feature increases as it is observed across a greater number of frames. Consequently, the triangulated point associated with the feature tends to be more precise.

As a result of the above discussion, it can be concluded that the lower $N_{f_{min}}$ is set, the higher both the update rate and the accuracy will be. To validate the above-mentioned claim in practice, we analyzed the performance of the FMSCKF under different values of $N_{f_{min}}$ using the first $10$ seconds of the EuRoC MAV dataset MH\_$01$ \citep{ref27}. We calculated the ratio of \textit{position error} to \textit{output rate} for different values of $N_{f_{min}}$. A small ratio corresponds to either a small position error, a high output rate, or both, all of which are desirable. As a result, the introduced measure can be used to compare the performance of the proposed algorithm for different values of the parameter $N_{f_{min}}$. The smaller the ration, the better the performance. The result is shown in Figure \ref{fig: nfmin}. In accordance with our assertion, it is obvious from Figure \ref{fig: nfmin} that when a higher $N_{f_{min}}$ is used, the error-to-speed ratio is higher.

The reason for choosing the value 8 for $N_{f_{min}}$ in our implementations is that in our algorithm, we use the well-known 8-point RANSAC algorithm for outlier rejection \citep{longuet1981computer}. To compute the fundamental matrix using this approach, we need at least 8 points. However, one can decide not to use this outlier rejection algorithm and choose a smaller value for $N_{f_{min}}$. According to Figure \ref{fig: nfmin}, this will result in an even higher accuracy and output rate.

\section{Conclusion}\label{section6}
In this paper, we aimed to address the challenge of fast and precise pose estimation (PE) for agile autonomous robots. We introduced an enhanced variant of the well-known Multi-State Constraint Kalman Filter (MSCKF), named Fast MSCKF (FMSCKF), designed to tackle the high computational cost associated with real-time implementation on resource-constrained robots. The FMSCKF leverages innovative feature marginalization and state pruning techniques to achieve computational efficiency, making it approximately six times faster than the standard MSCKF, all while delivering a substantial improvement of at least 20\% in final position estimation accuracy. In section \ref{section5}, we thoroughly discussed the differences between the MSCKF and the FMSCKF and highlighted the reasons why the FMSCKF outperforms the MSCKF. Our extensive evaluation of the FMSCKF on both public datasets and in real-world experiments demonstrated the remarkable performance of the FMSCKF compared to the state-of-the-art algorithms. Future works include exploring machine-learning-based feature extraction and tracking algorithms and evaluating their performance and robustness in VIO applications. Another direction could be using additional sensors, such as LiDAR, to enhance the performance of the proposed algorithm.

\bibliographystyle{elsarticle-harv} 
\bibliography{ref}

\end{document}